# A Review On Table Recognition Based On Deep Learning

Shi Jiyuan[1], Shi Chunqi[2]

[1]( Information hub of Hong Kong University of Science and Technology(Guangzhou),Guangzhou 511453,China )

[2]( China Pacific Insurance (Group) Co., Ltd.,Shanghai 200090,China )

**Abstract** Table recognition is using the computer to automatically understand the table, to detect the position of the table from the document or picture, and to correctly extract and identify the internal structure and content of the table. After earlier mainstream approaches based on heuristic rules and machine learning, the development of deep learning techniques has brought a new paradigm to this field. This review mainly discusses the table recognition problem from five aspects. The first part introduces data sets, benchmarks, and commonly used evaluation indicators. This section selects representative data sets, benchmarks, and evaluation indicators that are frequently used by researchers. The second part introduces the table recognition model. This survey introduces the development of the table recognition model, especially the table recognition model based on deep learning. It is generally accepted that table recognition is divided into two stages: table detection and table structure recognition. This section introduces the models that follow this paradigm (TD and TSR). The third part is the End-to-End method, this section introduces some scholars' attempts to use an end-to-end approach to solve the table recognition problem once and for all and the part are Data-centric methods, such as data augmentation, aligning benchmarks, and other methods. The fourth part is the data-centric approach, such as data enhancement, alignment benchmark, and so on. The fifth part summarizes and compares the experimental data in the field of form recognition, and analyzes the mainstream and more advantageous methods. Finally, this paper also discusses the possible development direction and trend of form processing in the future, to provide some ideas for researchers in the field of table recognition. (Resource will be released at https://github.com/Wa1den-jy/Topic-on-Table-Recognition .)

**Key words** table recognition; deep learning;table detection;table structure recognition;End-to-End;Data-centric

**摘要** 表格识别，即利用计算机自动处理表格，实现从文档或者图片中检测出表格位置，并对表格内部结构和内容进行正确的提取和识别。在经过早期基于启发式规则和基于机器学习的主流方法之后，深度学习技术的发展为这一领域带来了新的范式。本综述主体从五个方面探讨了表格识别问题。第一部分介绍数据集、基准和常用评价指标，本节选取研究人员使用频率较高且具有代表性的数据集、基准和评价指标。第二部分介绍表格识别模型，本文介绍表格识别模型的发展历程，着重介绍了基于深度学习的表格识别模型。目前普遍认为表格识别分为两个阶段:表格检测和表格结构识别，本节介绍遵循这一范式的模型(表格检测和表格结构识别)。第三部分是端到端方法，本节介绍了一些学者尝试使用端到端方法一次性地解决表格识别问题。第四部分是以数据为中心的方法，如数据增强、对齐基准等方法。第五部分是对表格识别领域的实验数据做了汇总与比较，分析了主流和较为有优势的方法。最后，本文还对未来表格处理可能的发展方向和趋势进行了探讨，以期为表格识别领域的研究人员提供一些思路。

表格在教科书、报纸、屏幕、可移植文档格式（Portable Document Format，PDF）等具有传递信息功能的媒介中很常见。对于上述媒介的读者来说，表格是信息高度精炼集中的形式，表格允许读者快速理解现有信息的同时给读者提供了进一步探索阅读内容的可能。

在当今这个信息化时代，读者只能从视觉上获取表格信息而无法将其自动准确得转化成可编辑的文本信息会给读者带来极大的不便。因而从20世纪末开始，人们就开始尝试使用计算机技术来正确检测文档或者图片中的表格位置以及识别表格内部结构和内容，而表格识别领域也慢慢进入研究者的视野，吸引了无数研究者的投入研究。

表格识别在科学界被认为是一个具有挑战性的课题。在这一领域已经进行了大量的研究，尽管大多数研究都有局限性[2,3]。对于研究者来说，由于表格类型和表格布局的多样性，因此创建一个通用的检测和提取文档中表格信息的模型是极为困难的。Hu等[1]将表格识别问题的处理分为了表格检测和识别两个子任务，表格检测简言之就是追求精准定位到表格的确切位置；而表格识别则是对表格内部结构进行分析，以求完整并且精准地复原整体表格原本的结构。而这种先检测后识别的两步骤也成为了后世研究表格问题的主流范式。但是不可忽视的是，也有一些学者将端到端的思想引入了表格识别领域，即将之前表格检测和表格结构识别两个子任务合二为一，一次性完成从输入端到输出端的转化。还有学者尝试以数据为中心的方向引入这个领域，都取得了长足的进展，本综述也会分别进行讨论。

从20世纪末表格识别问题被定义开始，研究者们首先尝试用基于规则或者基于启发式的方法解决这个问题。尽管取得了一定的成效，但是依旧存在着通用性不足，人力成本太高等诸多瓶颈。21世纪初的十年，由于机器学习技术的发展，学者们开始使用机器学习算法来检测和识别表格。基于机器学习算法的表格识别减少了人力成本的投入，提高了算法的通用性，但是表格结构识别正确率不算高，算法泛化性不够强依旧困扰着研究者们。近年来随着神经网络的进一步发展，深度学习技术的不断演进，使表格检测和结构识别技术发展得更加迅速。深度学习模型现在被广泛应用于表格检测和识别的许多方面，包括通用的表格检测 [4,5,6,7,8]。另一方面，表格结构受到的关注相对少一些，主要是表格逻辑结构典型特征的信息提取[9,10,11]。在促成表格检测算法快速进步的各种变量和举措中，深度卷积神经网络和GPU计算能力的发展占据了重要的位置。目前对于表格检测任务的处理已经较为成熟，其中有些模型能在特定大型数据集上实现100%的准确率，但是表格结构处理仍旧是一个值得探索的问题，我们也看到了更多研究者在这个子任务进行探索。

本综述聚焦于表格识别领域。将从表格数据集的数据组成和简要对比差异、考察重要的表格检测算法以及概述它们随着时间的变化、对表格结构识别方法的梳理与介绍、基于端到端的方法、以数据为中心的方法以及表格识别领域未来展望这几个方面展开。

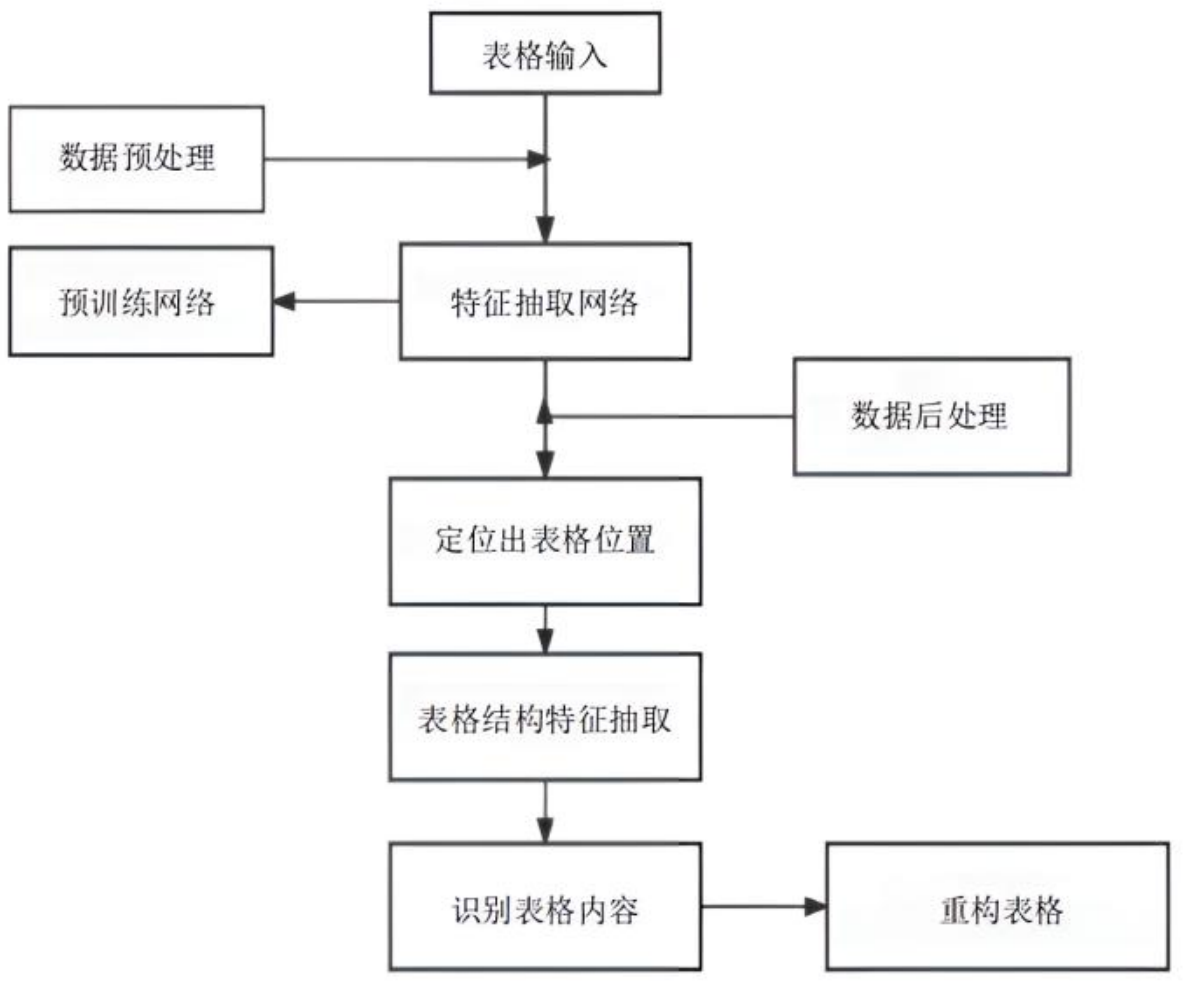

**图 1 使用深度学习处理表格的流程**

**Fig. 1 Deep Learning approaches for Table Undersatanding**

## 1 表格数据集、基线模型和常见评估指标

研究团队和研究者们为表格识别问题提供了极为丰富的数据集。这些数据集在用途上既面向表格检测，又面向表格结构识别；在分类上有的数据集专注于单一领域，如财经发票等，有的数据集则是包罗万象；在形式上有的数据集完全由电子文档（如HTML，LateX等）组成，而有的是由现实生活场景中的图片组成。总体而言，表格识别领域发展至今，数据集的多样化、复杂化的特点越来越切近于现实生活的方方面面。给研究者提供充足的研究资料的同时也带来了不同的挑战。

本综述归纳了其中较为经典的数据集给读者进行分析介绍。其中一些数据集不仅提供了单纯的数据资源，还给研究者们提供了基准模型或者做出了自己基于该数据集的实验数据，以给后来的研究者提供研究的思路和启发。

### 1.1 UW-3

UW-3[26]是1996年发布的数据集，其中包含1600张经过倾斜校正的英文文档图像，这些图像带有手动编辑的实体边界框。页面帧、文本和非文本区域、文本行和单词都被这些边界框包围。每个区域的类型（文字、数学、表格等）也都有所标明。数据集中大约有120个文档图像，至少有一个标记的表格区域。每个图像的实体边界框都存储在一个可扩展标记语言（eXtensible Markup Language，XML）中。

### 1.2 UNLV

UNLV[25]是2012年发布的数据集，其中包含了2889张从各种来源（杂志、报纸、商业信函、年度报告等）收集的扫描文档图像。这些扫描文档图像有双色调、灰度和传真格式可供选择，分辨率从200到300像素点不等。除了原始扫描文档图像数据集，还有实体边界框数据，其中包含手动标记的区域，区域类型

以文本格式提供。数据集中427个文档图像有至少一个标记好的区域。

**1.3 Marmot**

Marmot[24]是2012年发布的数据集，其被认为是表格检测领域的第一个大型数据集。它包含2000个带有实体边界框数据的可移植文档格式（Portable Document Format，PDF）页面。这项标注任务由 15 人完成。为了减少主观性，建立了统一的标注标准，对每个拥有实体边界框的页面都进行了复核。数据集中的电子文档页面具有广泛的语言类型、页面布局和表格样式。中文页面是从方正Apabi数字图书馆的120多本不同主题领域的电子书中挑选出来的，每本不超过15页。英文页则是从网上检索而来。从1970年到2011年，爬取了 1500 多篇会议和期刊论文，涵盖了广泛的主题。其中中文电子书页面大多是一栏，而英文页面则是一栏和两栏都有。这个数据集包含了广泛的表格类型，从直线表格到部分和非直线表格，从水平表格到垂直表格，从内列表格到跨列表格等。

**1.4 ICDAR 2013**

ICDAR 2013[21]国际文档分析与识别会议（International Conference on Document Analysis and Recognition，ICDAR）在2013年发布竞赛的官方实践数据集。作者并没有把重点放在一个特定的文档子集上，而是尽可能广泛地评估表格识别领域，而这个数据集，以及实际的比赛数据集，都是通过从谷歌搜索中系统地收集可移植文档格式（Portable Document Format，PDF）生成的，以便使选择尽可能客观。ICDAR 2013数据集包含150个表，其中75个来自27份欧盟文件摘录，75个来自40份美国政府文件摘录。表格区域是页面上的矩形区域，其坐标定义了它们。多个区域可以包含在同一个表中，因为一个表可以跨越多个页面。表格检测和表格结构识别是ICDAR 2013的两个子任务。

在ICDAR 2013举行的表格竞赛跨越几种输入格式。比赛包含三个任务:(i)表格位置检测，(ii)表格结构识别，以及(iii)前两个任务的结合。该竞赛收到了七个研究团队的结果，其中表格位置检测任务中最优的结果可以实现F1为0.9848，表格结构识别任务最优的结果是F1为0.9460而最后任务的最优F1的值是0.8772。

**1.5 ICDAR 2015**

ICDAR2015[21]国际文档分析与识别会议（International Conference on Document Analysis and Recognition，ICDAR）在2015年发布竞赛的官方实践数据集。该次官方给出的竞赛包含了定位、分割、识别和端到端的四个任务。其中任务四（端到端）包含1,670张图像(17548个注释区域)。而任务三（识别）的视频文本的数据集序列数量达到49个，包括27824帧(184687个注释区域)。

在ICDAR 2015举行的表格竞赛包含了定位、分割、识别和端到端的四个任务。其中任务一最优团队的F1得分是0.4984，而任务四最优的F1得分则是0.437。

**1.6 ICDAR 2017**

ICDAR 2017[22]国际文档分析与识别会议（International Conference on Document Analysis and Recognition，ICDAR）在2017年发布竞赛的官方实践数据集。与ICDAR 2013表数据集相比，该数据集明显更大。其中总共有2417张图片，包括图表、表格和公式。 数据集通常分为1600张图片，被分为了训练集和测试集，其中有731个表格区域用于训练，剩余的817张图片有350个表格区域用于测试。该数据集经常被用来测试不同的表格检测方法。

在ICDAR 2017举行的表格竞赛中，所有团队都以深度学习作为基本方法，然后结合不同的传统特征或方法来提高检测性能。其中最优团队在IoU阈值0.8下的所有页面对象检测中平均F1达到了0.898，mAP值达到0.805。

**1.7 DeepFigures**

DeepFigures[32]是2018年提出的大型数据集。其在没有人工辅助的情况下，为大量科学论文中的图形提取任务生成高质量的训练标签。作者通过使用来自两个大型科学论文网络集(PubMed和arXiv)补充数据，在可移植文档格式（Portable Document Format，PDF）中定位图像和标题来实现这一点。作者提供的结果数据集约550万个表格和数字诱导标签，以促进这项任务的现代数据驱动方法的发展，比之前最大的数字提取数据集大4000倍，标注的平均精度为96.8%。

作者使用该数据集来训练用于端到端图形检测的深度神经网络，其中LaTeX表格的准确率、召回率和F1指标达到了1.0、1.0和1.0；而可扩展标记语言（eXtensible Markup Language，XML）格式的表格的准确率、召回率和F1指标则是达到了0.97、0.91和0.94，此外生成的模型与以前的工作相比可以更容易地扩展到新领域。

**1.8 ICDAR 2019**

ICDAR 2019[23]国际文档分析与识别会议（International Conference on Document Analysis and Recognition，ICDAR）在2019年发布竞赛的官方实践数据集。官方提出了表格检测(TRACK A)和表格结构识别(TRACK B)两种数据集，数据集分别由现代文档和带有手绘表格和手写文本的存档文档组成。数据集包含1600张用于训练的图像和839张用于测试的图像。现代文档类型的数据集在TRACK A和B中包含600张用于训练的图像和340张用于测试的图像。人工类型包含1200张图像在TRACK A和B用于训练和499张图像用于测试。

在ICDAR 2017举行的表格竞赛中，所提交的方法已在现代和人工数据集上进行了评估。研究表明，最优的表检测方法的加权平均F1高于0.9。相反，表识别只能达到加权平均F1值为0.48。这表明有表格识别任务还有明显的改进空间。

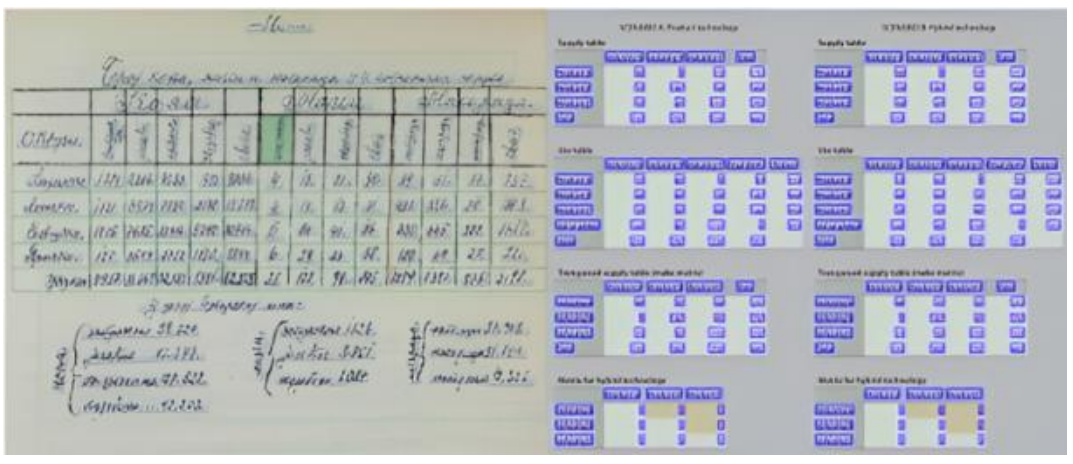

**图 2 ICDAR 2019 中的人工文档和现代文档[23]**

**Fig. 2 Human and modern documents in ICDAR 2019[23]**

## 1.9 SciTSR

SciTSR[28]是2019年发布的数据集。其中提供了一个大规模的表结构识别数据集，该数据集来自科学论文，包括来自可移植文档格式（Portable Document Format，PDF）文件的15000个表格，表格区域的图像，它们对应的结构标签和每个单元格的边界框。数据集被分成2000个训练数据和3000个测试数据。同时，还提供了一个名为SciTSR COMP的复杂表列表。这是一个LateX来源的数据集，并且是可移植文档格式（Portable Document Format，PDF）文件中用于表结构识别的第一个大规模数据集。

作者们提出的模型数据集上都优于最先进的基准模型。在ICDAR-2013和SciTSR数据集上，F1值比基准至少高出0.02，达到了0.819和0.936。而在SciTSR-COMP中，F1值上优于其他方法至少0.07，达到了0.934。

## 1.10 PubTabNet

PubTabNet[30]是2020年发布的数据集，也是最大的可公开访问的表识别的数据集之一，包括568k图片的结构化超文本标记语言（Hyper Text Markup Language，HTML）表示。尽管数据提供了表格结构和字符，但缺少实体边界框。PubTabNet是通过比较PubMed Central™开放存取子集中科学出版物的可扩展标记语言（eXtensible Markup Language，XML）和可移植文档格式（Portable Document Format，PDF）格式自动构建的。

作者们提出了一种新的基于注意力的编码器-双解码器架构，将表格图像转换为超文本标记语言（Hyper Text Markup Language，HTML）。该模型具有结构解码器，可以重构表结构，帮助单元格解码器识别单元格内容。此外，作者们还提出了新的基于树编辑距离的相似性（TEDS）度量用于表格识别，更适合确定有多行跨度的单元格不对齐和OCR误差的问题。实验表明，架构仅依靠图像表示就能准确识别复杂的表，其TEDS分数比目前最先进的方法高出0.097。

## 1.11 TableBank

TableBank[31]是2020年发布的数据集。其中提出了一种新的弱监督方法来自动创建数据集，这样生成的效果是比现有的用于表分析的人工标记数据集要大几个数量级。与传统的弱监督训练集不同，这种方法不仅可以生成大量而且高质量的训练数据。而且目前互联网上有很多电子文档，比如微软的Word和Latex文件。根据定义，这些在线文档在其源代码中包含表的标记标签。直观地说，这些源代码通过使用标记语言在每个文档中添加边界框进行操作。可以修改Word文档的Office XML代码，以识别每个表格的边界。Latex文档的代码也可以通过修改来识别表格边界框。这种方法可以为广泛的领域生成高质量的标记数据，包括商业文档，官方文件、研究论文等，这对于大规模的表分析任务非常有用。TableBank数据集由来自不同领域的417234个高质量标记表及其原始文档组成。

作者在表格检测任务中使用Faster R-CNN模型作为基线模型，此外，该模型通过共享RPN和Fast R-CNN的卷积特征，将它们合并为一个网络，这样网络就可以以端到端的方式进行训练。而在表格结构识别任务上利用图像到文本模型作为基线，使用Image-to-markup[78]作为基线模型，在TableBank数据集上训练模型。

## 1.12 FinTab

Fintab[29]是2021年发布的数据集。该数据集一个标准中文数据集，专注于财经领域，其中包含1600多种不同类型的财务表及其在JSON中的相应结构表示形式。作者们总共收集了19个可移植文档格式（Portable Document Format，PDF）文件，所有文档加起来有3329页，其中2522页包含表格。FinTab是手动审核的，这使得其更加可靠。作者还提供字符和字符串作为文本基础真相，给出了表的单元格和表线的详细信息。

作者们还提出了一种新的基于图神经网络的算法GFTE来完成表结构识别任务，作为一个基线模型。其中横轴预测准确率可以达到0.8610，而纵轴预测准确率可以达到0.9030。

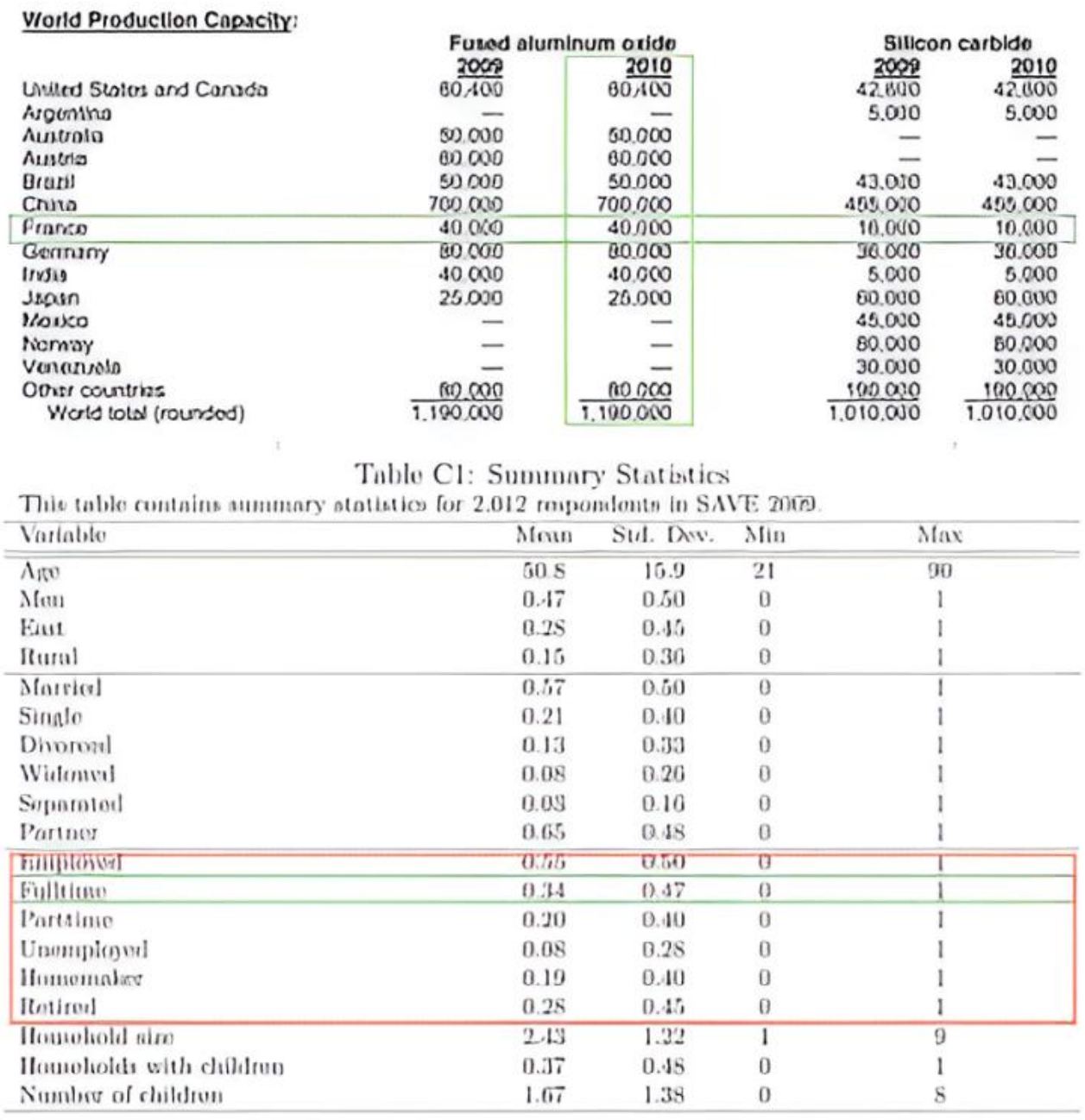

World Production Capacity:

| | Fused aluminum oxide 2009 | Fused aluminum oxide 2010 | Silicon carbide 2009 | Silicon carbide 2010 |
|---|---|---|---|---|
| United States and Canada | 60,400 | 60,400 | 42,600 | 42,600 |
| Argentina | — | — | 5,000 | 5,000 |
| Australia | 50,000 | 50,000 | — | — |
| Austria | 60,000 | 60,000 | — | — |
| Brazil | 50,000 | 50,000 | 43,000 | 43,000 |
| China | 700,000 | 700,000 | 455,000 | 455,000 |
| France | 40,000 | 40,000 | 10,000 | 10,000 |
| Germany | 80,000 | 80,000 | 30,000 | 30,000 |
| India | 40,000 | 40,000 | 5,000 | 5,000 |
| Japan | 25,000 | 25,000 | 60,000 | 60,000 |
| Mexico | — | — | 45,000 | 45,000 |
| Norway | — | — | 80,000 | 80,000 |
| Venezuela | — | — | 30,000 | 30,000 |
| Other countries | 80,000 | 80,000 | 100,000 | 100,000 |
| World total (rounded) | 1,190,000 | 1,190,000 | 1,010,000 | 1,010,000 |

Table C1: Summary Statistics

This table contains summary statistics for 2,012 respondents in SAVE 2009.

| Variable | Mean | Std. Dev. | Min | Max |
|---|---|---|---|---|
| Age | 50.8 | 15.9 | 21 | 90 |
| Men | 0.47 | 0.50 | 0 | 1 |
| East | 0.28 | 0.45 | 0 | 1 |
| Rural | 0.15 | 0.36 | 0 | 1 |
| Married | 0.57 | 0.50 | 0 | 1 |
| Single | 0.21 | 0.40 | 0 | 1 |
| Divorced | 0.13 | 0.33 | 0 | 1 |
| Widowed | 0.08 | 0.26 | 0 | 1 |
| Separated | 0.03 | 0.16 | 0 | 1 |
| Partner | 0.65 | 0.48 | 0 | 1 |
| Employed | 0.55 | 0.50 | 0 | 1 |
| Fulltime | 0.34 | 0.47 | 0 | 1 |
| Parttime | 0.20 | 0.40 | 0 | 1 |
| Unemployed | 0.08 | 0.28 | 0 | 1 |
| Homemaker | 0.19 | 0.40 | 0 | 1 |
| Retired | 0.28 | 0.45 | 0 | 1 |
| Household size | 2.43 | 1.32 | 1 | 9 |
| Households with children | 0.37 | 0.48 | 0 | 1 |
| Number of children | 1.67 | 1.38 | 0 | 8 |

**图 3 FinTab 中具有挑战性的表格结构识别数据[29]**

**Fig. 3 Challenging table recognition data in FinTab[29]**

## 1.13 PubTables-1M

PubTables-1M[27]是2022年发布的数据集，其中包含了近100万个来自科学文章的表格，支持多种输入方式，并包含了表结构的详细标题和位置信息，这使得它对各种建模方法都很有用。它还使用一种新的规范化过程，解决了在之前的数据集中观察到的被称为过度分割的实体边界框不一致的重要来源。

作者们还展示了在PubTables-1M上训练的Transfromer-based的对象检测模型，模型对检测、结构识别和功能分析这三个任务都产生了出色的结果，而不需要对这些任务进行任何特殊定制。

| | | ΔSDM | | | |
|---|---|---|---|---|---|
| | | better | equal | Worse | Sum |
| ΔSCA | better | 19457 (28.9) | 12 (0.02) | 14654 (21.8) | **34,123 (50.8)** |
| | equal | 1158 (1.7) | 21989 (32.7) | 1024 (1.5) | **24,171 (36.0)** |
| | worse | 3755 (5.6) | 2 (0.003) | 5183 (7.7) | **8,940 (13.2)** |
| | Sum | 24370 (36.2) | 22003 (32.7) | 20861 (31.0) | 67,234 (100.0) |

(a) Oversegmented structure annotation

| | | ΔSDM | | | |
|---|---|---|---|---|---|
| | | better | equal | Worse | Sum |
| ΔSCA | better | 19457 (28.9) | 12 (0.02) | 14654 (21.8) | **34,123 (50.8)** |
| | equal | 1158 (1.7) | 21989 (32.7) | 1024 (1.5) | **24,171 (36.0)** |
| | worse | 3755 (5.6) | 2 (0.003) | 5183 (7.7) | **8,940 (13.2)** |
| | Sum | 24370 (36.2) | 22003 (32.7) | 20861 (31.0) | 67,234 (100.0) |

(b) Canonical structure annotation

**图 4 过度注释和合理注释的表格[27]**

**Fig. 4 Overcommented and reasonably commented tables[27]**

## 1.14 TabRecSet

TabRecSet[76]是2022年发布的数据集，该数据集是端到端表格识别的最大和第一个双语数据集，有38.1K个表格，其中20.4K为英语，17.7 K为中文。数据有多种形式，如边界完整和不完整的表格，规则和不规则的表格(旋转，扭曲等)。多种多样的现实场景，从扫描图像到相机拍摄的图像，从文档到电子表格，从教育试卷到财务发票。注释分别包括表格检测、表格结构识别和表格内容识别的表体空间注释、单元格空间和逻辑注释以及文本内容。空间标注采用多边形，而不是大多数数据集采用的边界框或四边形。其中多边形空间注释更适合于在野外场景中常见的不规则表。此外，为了提高表格标注的效率和质量，作者们还提出了一个可视化的交互式标注工具TableMe。

## 1.15 WikiTableSet

WikiTableSet[77]是2023年发布的数据集，是公开的基于图像的表识别数据集，有三种语言。其中包含近400万张英文表格图像、590K张日文表格图像和640K张法文表格图像，并带有相应的HTML表示和单元格边界框。

作者提出了一个弱监督模型WSTabNet模型用于表格识别，该模型仅依赖于表图像的HTML(或LaTeX)代码级注释。该模型由三个主要部分组成:用于特征提取的编码器、用于生成表结构的结构解码器和用于预测表中每个单元内容的单元解码器。我们的系统通过随机梯度下降算法进行端到端的训练，只需要表图像及其基本真实的HTML(或LaTeX)表示。

| | Lactic acidosis cases | Non-cases |
|---|---|---|
| *n* | 82 | 2510 |
| Age (years) | 73.5 (9.5) | 73.0 (11.5) |
| Male (%) | 65.8% | 54.6% |
| Lactate (mmol/L) | 9.3 (3.5) | 2.3 (1.7) |
| Bicarbonate (mmol/L) | 11.9 (3.9) | 23.5 (5.0) |
| Type 2 Diabetes, *n* (% of cohort) | 70 (85.4%) | 1676 (66.8%) |
| Users of metformin | 42 (60%) | 970 (57.9%) |
| Non-users of metformin | 28 (40%) | 706 (42.1%) |
| AKI stages, *n* (% of cohort with each stage of AKI) | | |
| Stage 0 | 17 (20.7%) | 1569 (62.5%) |
| Stage 1 | 16 (19.5%) | 508 (20.2%) |
| Stage 2 | 25 (30.5%) | 266 (10.6) |
| Stage 3 | 24 (29.3%) | 167 (6.6%) |
| Baseline eGFR, *n* (% of cohort) | | |
| >60 mL/min/1.73 m$^2$ | 59 (71.9%) | 1979 (78.8%) |
| 30-59 mL/min/1.73 m$^2$ | 19 (23.2%) | 468 (18.6%) |
| <30 mL/min/1.73 m$^2$ | 4 (4.9%) | 63 (2.5%) |

**图 5 边界框很少（弱监督）的表格[77]**

**Fig. 5 Table with few bounding boxes (weakly supervised)[77]**

## 1.14 常见评价指标

表格检测模型往往使用多个指标来衡量模型的性能，即每秒帧数(Frames Per Second，FPS)，精度（Average Precision，AP）和召回率（Average Recall，AR）。一般而言，所有类别的平均精度的平均值，称为精度均值(mAP)，精度均值(mAP)是最常见的评估指标。精度的计算公式是交集面积/联合面积(Intersection over Union，IoU)，即基实体边界框与预测边界框的重叠面积与联合面积之比。设置一个阈值来确定检测是否正确。如果 IoU 超过阈值，则将其归类为真正(True Positive，TP)，而低于该阈值的IoU则归类为假正（False Positive，FP）。如果模型未能检测到实体边界框中存在的对象，则被称为假阴性。精度衡量正确预测的百分比，而召回率衡量相对于实体边界框的正确预测。

$$\text{Average Precision(AP)} = \frac{True\ Positive\ (TP)}{(True\ Positive\ (TP) + False\ Positive\ (FP))} \tag{1}$$

$$\text{Average Recall(AR)} = \frac{True\ Positive\ (TP)}{(True\ Positive\ (TP) + False\ Negative\ (FN)\ )} \tag{2}$$

$$\text{F1}-\text{score} = \frac{2*(AR*AP)}{(AR+AP)} \tag{3}$$

IoU是一个用来发现实体边界框和预测表格边框之间差异的指标。这个指标被用于大多数流行的对象检测算法。在对象检测中，模型为每个对象预测多个边界框，并根据每个边界框的置信度分数，根据它们的阈值删除不必要的边界框。我们需要根据我们的要求声明阈值。

$$\text{IoU} = \frac{Area\ of\ union}{area\ of\ intersection} \tag{4}$$

除了以上最为常见的评价指标之外，Zhong X等[30]认为计算精度、召回率和F1值等指标存在两个问题：1）这些指标只检测非空单元格之间的邻接关系，无法检测出空单元格和超出近邻单元格的不对齐引起的误差；2）由于上述指标通过精确匹配检查关系，导致没有一个机制来衡量细粒度的单元格内容识别性能。因此作者提出一种新的评价指标：基于树-编辑-距离的相似度（Tree-Edit-Distance-based Similarity，TEDS）。该指标解决了通过检查全局结构的识别结果，两个单元格的相同，方向匹配的水平，允许它识别所有类型的结构错误；此外计算树编辑操作为节点替换时的字符串编辑距离。这样的指标比上述指标更合适地捕获跨越多行的单元格偏差和OCR误差。

在使用TEDS度量时，我们需要以HTML格式的树形结构表示表。最后，计算两个表格之间的TEDS为：

$$\text{TEDS(Ta,Tb)} = 1 - \frac{\text{EditDist(Ta, Tb)}}{\max(|\text{Ta}|, |\text{Tb}|)} \tag{5}$$

其中Ta和Tb是HTML格式表格的树状结构。EditDist表示树编辑距离，|T|表示树中的节点数

| Dataset | #Images | #Tables | TSR | Baseline/ Experiment | Year | #Annotations |
|---|---|---|---|---|---|---|
| UW-III | 120 | 120 | √ | × | 1996 | |
| UNLV | 427 | 558 | √ | × | 2012 | |
| Marmot | 2000 | 2000 | √ | × | 2012 | |
| ICDAR 2013 | 128 | 156 | √ | √ | 2013 | 竞赛数据集 |
| ICDAR 2015 | 1670 | 1670 | √ | √ | 2015 | 竞赛数据集 |
| ICDAR 2017 | 2417 | 1020 | × | √ | 2017 | 竞赛数据集 |
| DeepFigures | 1.67M | 1.4M | × | √ | 2018 | |
| ICDAR 2019 | 2439 | 3.6K | √ | √ | 2019 | 竞赛数据集 |
| SciTSR | 15K | 15K | √ | √ | 2019 | |
| PubTabNet | 568K | 568K | √ | √ | 2020 | |
| TableBank | 278K | 417K | √ | √ | 2020 | |
| FinTab | 1685 | 1685 | √ | √ | 2021 | 财经数据集 |
| PubTables-1M | 1M+ | 1M+ | √ | √ | 2022 | |
| TabRecSet | 32.07K | 38.17K | √ | × | 2022 | 端到端数据集 |
| WSTabNet | 5M | 5M | √ | √ | 2023 | |

**表 1 较为常见的表格数据集**

**Table. 1 More common table dataset**

## 2 表格识别模型

表格处理是文本分析中的一个关键任务，有助于从非结构化或半结构化文档中提取结构化数据。目前的研究者普遍将表格处理分为三个步骤:表格检测、表结构识别以及内容提取和解释。表格检测和表格识别由于其对表格本身性质的分析和处理而被统一划分到表格识别这个领域中，Hu[1]等人将表格识别任务分为了表格检测和结构识别两个子任务。

从上世纪研究者们就开始尝试将计算机技术引入表格识别领域中来，其中从1990年到2000年这十年间主流研究是基于启发式/规则式的方法，而后的十年随着一系列经典机器学习算法的流行，研究者们将机器学习的方法引入到表格识别中来。而随着深度学习的兴起，尤其从2016开始，深度学习和人工智能给表格识别领域带来了新的范式。

本段将首先在2.1节介绍基于启发式/规则式的表格识别方法，其次在2.2节介绍基于机器学习的表格识别方法。而2.3节会着重介绍基于深度学习的表格识别模型。

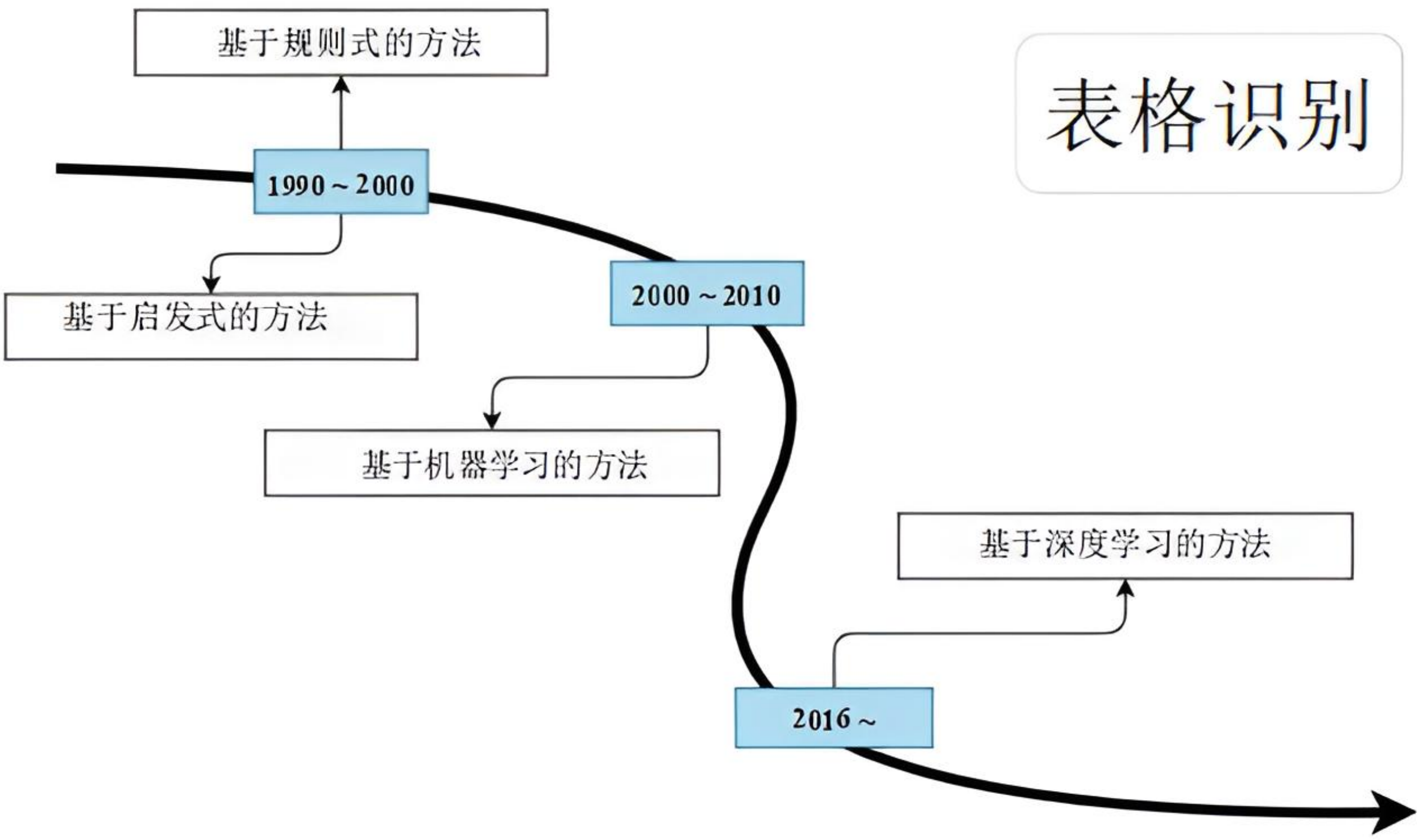

**图 6 表格识别领域随时间发展历程**

**Fig. 6 The evolution of the field of table recognition over time**

## 2.1 基于启发式/规则式方法

基于启发式的表格识别模型在20世纪90年代到21世纪10年代之间较为流行，研究者们采用不同的视觉线索，如图表分隔、图表布局结构以及空间特征等来检测表格。

Pyreddy等[33]提出了一种使用字符对齐、孔和间隙来检测表格的方法。Wang等[34]使用统计方法根据连续单词之间的距离来检测表行。采用分组水平连续词与垂直相邻行一起找出表格候选。Jahan等[35]提出了一种利用字间距和行高局部阈值检测表格区域的方法。Itonori等[36]提出了一种基于规则的方法，通过文本块排列和规则行位置来定位文档中的表格。Chandran等[37]发展了另一种基于垂直线和水平线的表格检测方法。Hassan等[38]通过分析文本块的空间特征来定位和分割表格结构。

Ruffolo等[39]提出了PDF-trex，这是一种启发式自底向上方法，用于单列可移植文档格式（Portable Document Format，PDF）中的表格识别。它使用页面元素的空间特征来对齐并将它们分组到段落和表格中。Nurminen等[40]提出了一组启发式方法来定位具有共同对齐方式的后续文本框，并为它们分配成为表格的概率。Fang等[41]使用表头作为起点来检测表区域并分解其元素。Harit等[42]提出了一种基于识别独特的表开始和表结束模式的表检测技术。上述方法在统一布局的文档中效果相对较好。然而，启发式规则需要调整到更广泛的表，并不真正适合通用解决方案。因此，机器学习方法开始被用于解决表检测问题。

## 2.2 基于机器学习方法

基于机器学习的表格检测模型则是集中出现于21世纪00年代到21世纪10年代这十年间，这个阶段的学者们会采用一些经典的机器学习算法，如应用XY树监督学习，应用KNN算法等经典机器学习模型对表格进行检测。

Kieninger等[43]采用了一种无监督的聚类分词学习方法来对表格进行检测。Cesarini等[44]使用了一种改进的XY树监督学习方法。Fan等[45]使用监督和非监督两种方法对可移植文档格式（Portable Document Format，PDF）中的表进行检测。Wang等[46]将决策树和SVM分类器应用于布局、内容类型和词组特征。T. Kasar等[47]使用连接检测，然后将信息传递给支持向量机分类器。Silva等[48]将联合概率分布应用于视觉页面元素的连续观测(隐马尔可夫模型)，将潜在的表行合并为表。Klampfl等[49]比较了两种来自数字科学文章的无监督表识别方法。Docstrum算法[50]利用KNN将结构聚合成行，然后利用行与行之间的垂直距离和夹角将它们组合成文本块。必须指出的是，这个算法

是设计于1993年，比本节提到的其他方法要早。F Shafait等[51]提出了一种有用的表识别方法，该方法在具有一系列布局的文档(包括商业报告、新闻故事和杂志页面)上表现良好。之后的Tesseract OCR引擎提供了该算法的开源实现。在机器学习经典算法的引入和应用下，表格识别领域得到了发展，但是很快，深度学习就以极大的优势几近席卷了整个领域，也给表格识别带来了新的范式。

## 2.3 基于深度学习方法

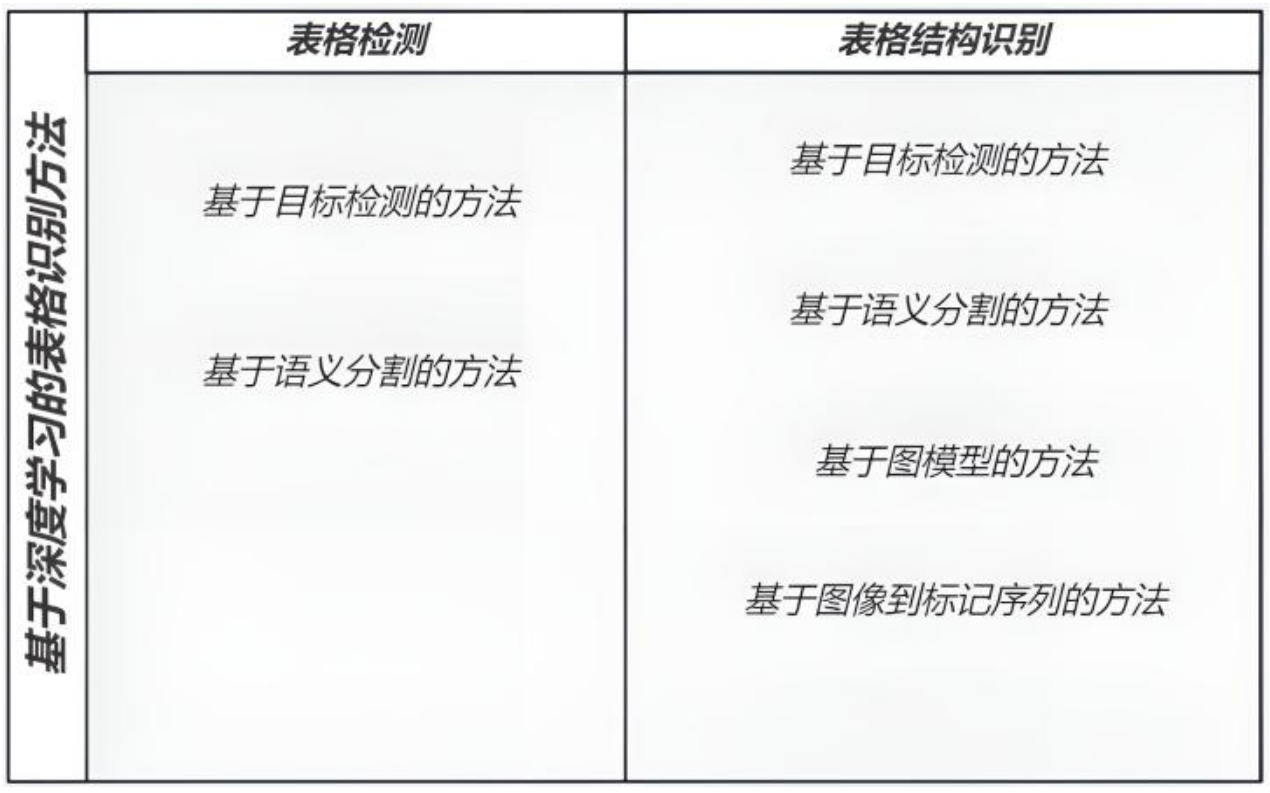

**图 7 基于深度学习的表格识别方法分类**

**Fig. 7 Table recognition method classification based on deep learning**

本节主要着眼于基于深度学习方法的表格识别模型。随着深度学习的兴起，基于深度学习的表格识别模型渐渐展示出了其优势，将不同神经网络模型运用于表格检测问题，概括地说，其优势在于以下几个方面：首先深度学习具有高精度和强大的表示能力，在表格检测问题上能够有效处理复杂的数据，如表格中各式各样的数据特征；其次深度学习的特征通常是自动学习的，不需要人工标注信息，同时它能够准确地从不同的数据中学习新特征，能够更好地处理一些模糊，复杂和非线性的数据；最后深度学习能够捕获表格的复杂结构信息，并能够从图像获得更多更有用的位置特征。

在深度学习模型被引入到表格识别领域中之时，研究者们往往会从表格检测和表格结构识别两个独立子任务着手去解决问题。伴随着深度学习和人工智能发展所催生出的计算机视觉和自然语言处理两个大方向，研究者们也对其中思想进行改进，进一步实验到表格识别问题上，在表格识别领域产生了极好的效果。其中在表格检测任务中，较为常见的方法时基于目标检测的方法和基于语义分割的方法；在表格结构识别中，基于目标检测的方法，基于语义分割的方法，基于图模型的方法和将图片转化为标记序列的方法都有研究者涉猎。

### 2.3.1 基于深度学习的表格检测

研究者在使用深度学习方法进行表格检测时，往往会借鉴计算机视觉领域的前沿技术和方法。在表格检测这个子任务上，目前主流研究者的方法分为两派：目标检测和语义分割。

#### 2.3.1.1 目标检测方法

从文档图像中检测表格可以表示为目标检测任务，将表格视为自然对象。表格检测中的目标检测涉及到识别文档图像中的表格结构。目标是将表格与其他文本或图形元素区分开来，并准确地确定它们的边界。

Hao等[8]和Yi等[96]应用R-CNN来检测表，其中Hao等[8]在提出的方法中，首先通过一些松散的规则，然后建立卷积网络并进行细化以确定所选区域是否为表是否。此外，表格区域的视觉特征直接通过卷积网络提取和利用，同时原始可移植文档格式（Portable Document Format，PDF）中包含的非视觉信息（例如字符、渲染说明）也被考虑在内，帮助实现更好的识别结果。而Yi等[96]在传统的基于卷积神经网络(CNN)的目标检测方法的基础上，重新设计了区域建议方法、训练策略和网络结构，并用动态规划算法取代了非最大抑制算法(NMS)。但这些方法的性能仍然依赖于启发式规则，在一些特定场景下需要对模块进行特殊适应性的调整，本质上说，就像以前的方法一样。

随后，更高效的单级目标检测器(如RetinaNet[97]和YOLO[98])和两级目标检测器(如Fast R-CNN[99]、Faster R-CNN、Mask R-CNN[100]和Cascade Mask R-CNN[101])被应用于文档图像中的图形和公式检测等其他文档对象。其中Nguyen D. Vo[55]等人将Fast RCNN和Faster RCNN的检测模型结合起来，扩展了Faster R-CNN的边界框，以增加目标检测的数量。将Fast R-CNN的有效边界框添加到Faster R-CNN的边界框中。N. Sun等[16]提出一种结合拐角点定位表格的方法，基于Faster R-CNN进行表格检测。作者提出的方法可以分两个步骤来实现。首先，表检测是通过Faster R-CNN实现。其次，通过可靠的拐角来细化表格边界。与以前的基于边界框的方法相比，该模型表检测的精度有显著提高。

A. Gilani等[13]提出的方法为了解决表格的不同布局和编码的问题，首先对文档图像进行预处理。然后将这些图像反馈送到区域建议网络，然后由全连接的神经网络进行表检测。作者所提出的方法在不同布局的文档图像(包括文档、研究论文和杂志)上都具有

高精度。Y Huang等[17]提出的基于YOLO原理的表格检测算法。作者基于视觉检测模型YOLO-V3并进行了各种适应表格检测任务的改进，包括一种锚点优化技术和两种后处理方法，以及采用k-means聚类进行寻找锚点，以解释文档对象和真实对象之间的显著差异。Madhav Agarwal等[79]提出了CDeC-Net模型，该模型由Mask R-CNN的多级扩展组成，具有可变形卷积的双主干，用于检测尺度变化的表，这样的方法使得模型在更高的IoU阈值下具有较高的检测精度。Saman Arif等[102]提出的解决方案是基于目标到表格往往包含更多的数字数据，因此应用颜色编码/颜色作为区分数字和文本数据的信号。此外作者还利用Faster R-CNN用于从文档图像中检测表格区域。此外值得一提的是，上面提到的方法中，[13,79,102]应用了不同的图像变换方法，如着色和扩张，以进一步改善结果。

此后Siddiqui等[58]将Deformation-convolution和RoI-pooling[103]结合到Faster R-CNN中，为几何修改提供了更高效的网络。Agarwal等[79]采用复合网络[104]作为主干，采用可变形卷积来提高两级级联R-CNN的性能。这些基于CNN的目标检测器有几个启发式阶段，如提案生成步骤和后处理步骤，如非最大抑制(NMS)。Tahira等[112]提出的半监督方法将检测视为一组预测任务，消除了锚点生成和后处理阶段(如NMS)，并提供了更简单，更有效的检测管道。

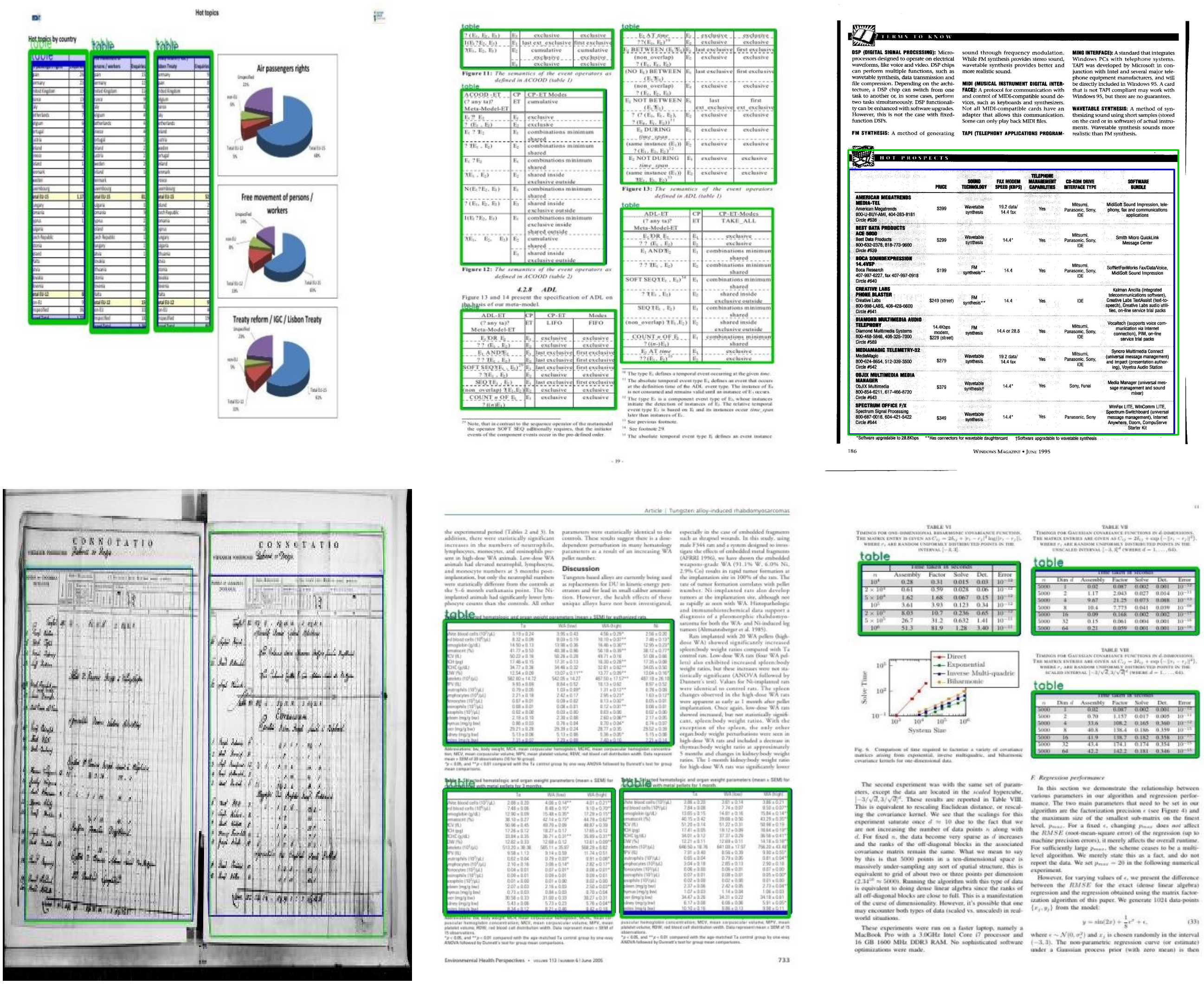

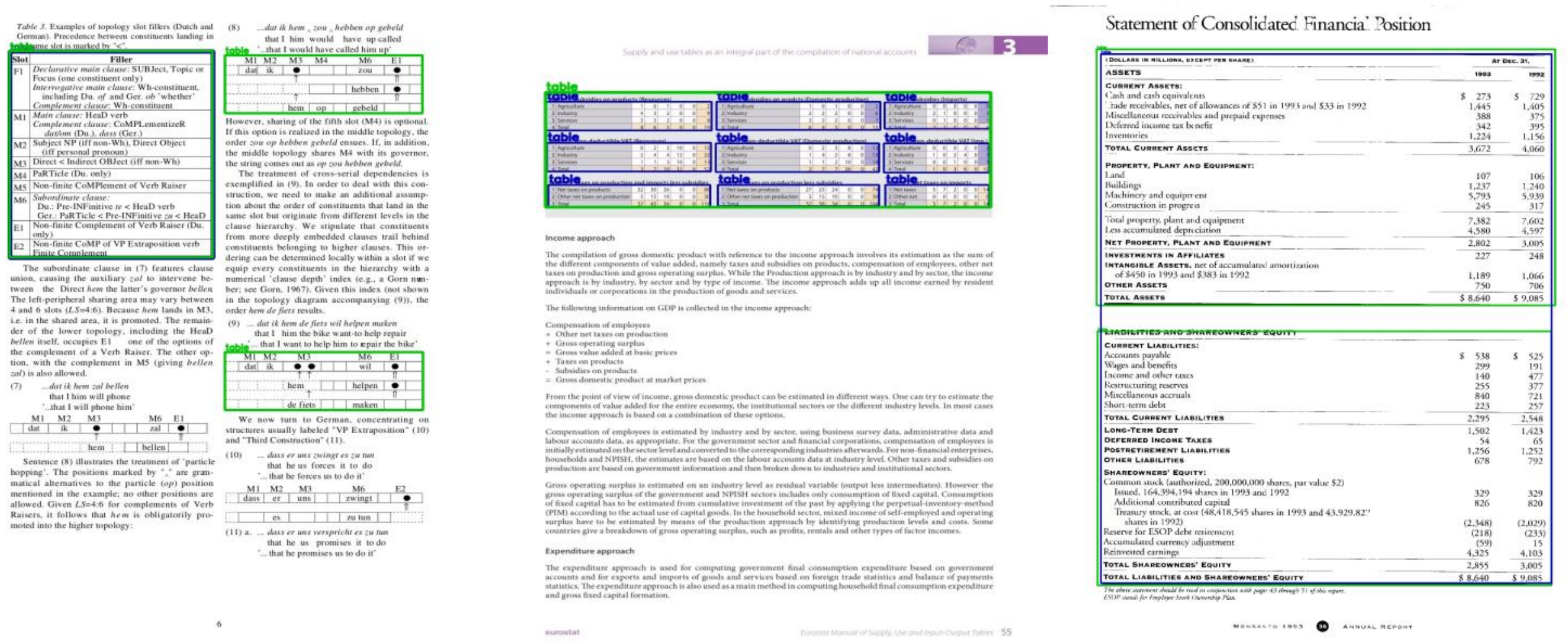

**图 8 基于目标检测方法的表格检测示意图[79]**

**Fig. 8 Table detection diagram based on object detection method[79]**

#### 2.3.1.2 语义分割方法

在表格检测的任务中，语义分割的方法也被许多研究者采用。基于语义分割的表格检测方法是文档分析和计算机视觉领域发展融合的一部分，是指对文档图像中的每个像素进行分类的过程，其中一个类别是“表格”。语义分割涉及到将文档图像划分为段的过程，每个段代表不同的语义类别，如文本，表格，图形等。与目标分割任务不同，表区域的边界不清晰和视觉线索稀疏限制了这些方法寻找准确表段的能力。但是这种方法对于从复杂文档中识别和提取表往往特别有效。

Yang等[105]提出了一种用于页面对象分割的全卷积网络，该网络结合了语言和视觉特征，以增强表和其他页面对象检测的分割结果。He等[106]提出了一种多尺度全卷积网络，它提供了分割掩码表/文本区域及其相关轮廓，然后进行细化以获得最终的表格块。I Kavasidis等[52]提出了一种结合DCNNs、图形模型和显著性（Saliency-based）思想来检测表格的方法。作者提出了一种用于表格检测的全卷积神经网络模型，对视觉线索执行多尺度推理。用于定位在数字化文档中定位表格。

### 2.3.2 表格结构识别

研究者在使用深度学习方法进行表格结构识别任务时，因为主体研究对象也是表格，因而主流研究方法与表格检测有一定程度上的重合，其中表格结构识别任务中也有基于目标检测方法和基于语义分割方法。此外还有基于图模型的方法和基于将图片转化为标记序列的方法。但是不同于表格检测任务的是，表格结构识别的任务还可以按照侧重点不同按照自顶向下（Top-down）的处理方式和自底而上(Bottom-up)的处理方式进行区分。

#### 2.3.2.1 目标检测方法

表结构识别领域中基于目标检测的方法侧重于识别和描绘文档中表的结构成分。基于目标检测的方法将表分解为行/列或单元格等较低层结构单元，并将其作为检测目标。目标是检测和分类表的各个结构元素，例如行、列、标题和单元格等组成元素，最后再合并成原本的表格结构。

SA Siddiqui等[58]提出的DeepTabStR基于可变形卷积网络，使用了一种独特的方法来分析文档中的表格图片。将表分隔成行/列，然后交叉行/列检测结果得到单元格边界框。但是，此方法不能处理交叉行和交叉列的情况。因此，X Zheng等[70]提出了全局表提取器(GTE)，通过联合检测表和识别单元结构的方法，可以在任何目标检测模型上实现结构识别。此外作者还开发了GTE-Table，基于表的固有单元格限制限制引入了一种新的惩罚机制。一种名为GTE-Cell的新型分层结构识别网络利用了表格结构。然而，上述方法假设表格对齐良好，并且检测到的单元格边界框都是矩形，这些都不适合自然场景表格。

针对自然场景下的表格问题，R. Long等[107]提出的Cycle-CenterNet包含一个循环配对模块，用于预测四边形边界框。在自然场景下取得了优异的效果。然而，四边形边界盒仍然难以准确地描述弯曲单元。

#### 2.3.2.2 语义分割方法

表结构识别领域中中基于语义分割的方法包括将图像中的每个像素划分为预定义的类别。在表结构识别的上下文中，这些类别对应于表的不同部分，如单元格、行、列、标题等。语义分割方法的主要任务是在粒度级别上准确描述和理解表的布局，确保正确识别和分类表的每个组件。这种方法在处理文档中的复杂表格时特别有效。

Sebastian Schreiber等[4]提出的DeepdeSRT首次采用Faster R-CNN对表格检测和结构识别，随后再基于语义分割进行表结构识别和SS Paliwal等[54]提出的TableNet对行和列进行语义分割，将行和列分割结果交叉得到单元格分割。

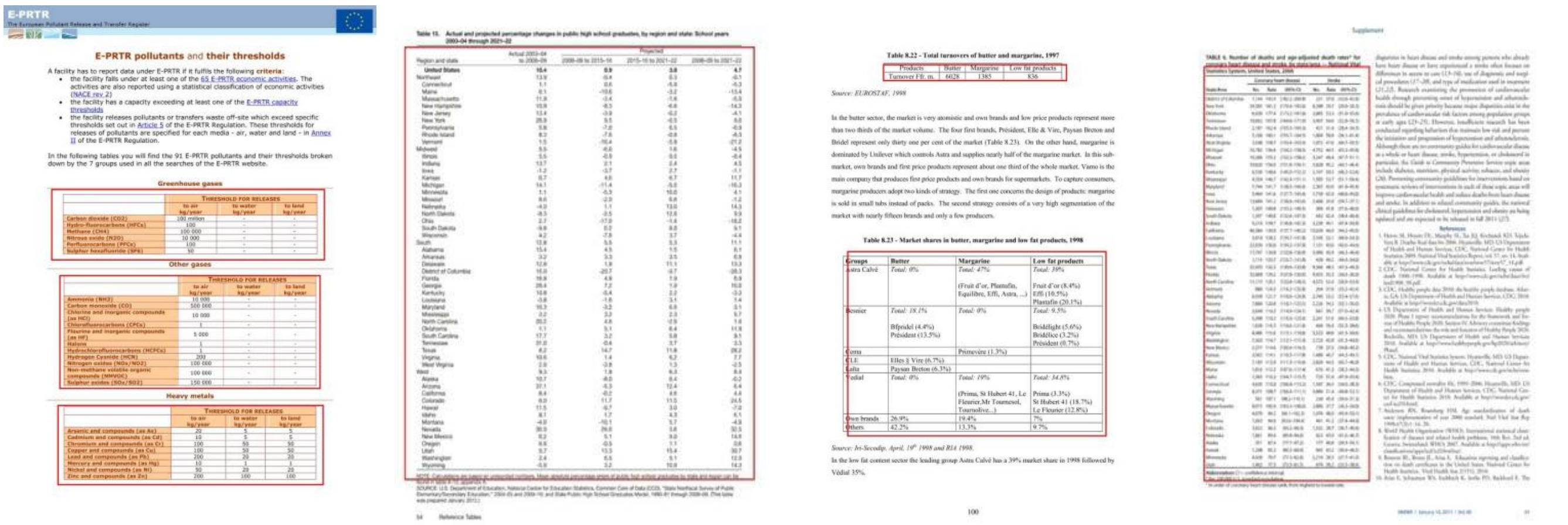

**图 9 DeepdeSRT 首次采用 Faster R-CNN 对表格检测[4]**

**Fig. 9 DeepdeSRT used Faster R-CNN for the first time to detect tables[4]**

SA Siddiqui等[58]将结构识别问题描述为语义分割问题。为了分割行和列，作者采用了全卷积网络。引入了预测分块的方法，它降低了表结构识别的复杂性，假设表结构中的一致性。作者从ImageNet导入了预训练模型，并使用了FCN的编码器和解码器的结构模型。该模型在给定图像时创建与原始输入图像相同大小的特征。

Z. Zhang等[71]提出的Split，Embed，and Merge (SEM)表结构识别模型。由拆分器、嵌入器和合并器三部分组成。在第一阶段，作者应用拆分器对表格行/列分隔符的潜在区域进行预测，得到表格的精细网格结构。在第二阶段中，通过充分考虑表中的文本信息，作者融合了视觉和文本形态的每个表格网格的输出特征。通过提供额外的文本特征。最后，作者以自回归的方式处理这些基本表网格的合并。相应的合并结果为通过注意力机制学习。C. Tensmeyer等[62]提出的SPLERGE (Split and Merge)模型使用扩张卷积的方法。使用两个不同的深度学习模型，第一个模型建立表格的网格状布局，第二个模型确定是否可以进一步跨越许多行或列的单元格。两个模型都使用段模型和合并模型划分单元格边界，并在交叉行/列情况下合并单元格。

上面提到方法处理都是从不同粒度(行/列，文本区域)的元素开始，这在某种程度上陷入了有损启发式规则或忽略空单元格分裂的问题。Qiao L等[108]提出的LGMPA通过充分利用所提出的局部特征中的文本区域和全局特征中的单元关系的视觉信息来获得更可靠的对齐边界框。具体而言，作者提出了局部和全局金字塔掩码对齐框架，该框架在局部和全局特征映射中都采用了软金字塔掩码学习机制。它允许边界框的预测边界突破了原始建议的限制。然后集成金字塔掩码重新评分模块，以折衷局部和全局信息并细化预测边界。最后，作者提出了一种具有鲁棒性的表结构恢复管道来获得最终结构，有效地解决了空单元格的定位和分裂问题。

在使用语义分割进行表格结构识别时，与对齐良好的表格相反，自然场景表倾向于倾斜、旋转和弯曲，从而破坏表的结构。Hongyi Wang等[114]提出了一种新的分割协同对齐网络(SCAN)。SCAN通过分割协作模块将表格单元格的位置和逻辑信息结合起来，精确分割单元格区域。其中单元格对齐模块对分割结果进行对齐，以恢复扭曲的表格结构。

#### 2.3.2.3 图模型方法

基于图的表格结构识别方法顾名思义将表格表示为图结构。图节点表示每个表格单元格的内容、表格图像中的嵌入向量或表格单元格的几何坐标。图的边定义了节点之间的关系，例如，它们属于同一列、行或表格单元格。这种方法特别适合于理解复杂多变的表布局。基于图模型的表格结构识别方法主要目标

是准确地建模和解释表中的结构关系，例如行和列的结构，以及标题和数据单元格的层次结构。

E Koci等[57]在确定每个单元格的布局角色后构建布局区域。我们对空间相互关系进行编码，在这些区域之间使用图形表示。在此基础上，他们提出了基于一组精心挑选的标准表识别算法 RAC (Remove and Conquer)。SR Qasim等[60]提出了两种基于图神经网络的梯度表格识别架构，作为典型神经网络的优越替代方案。作者提出的架构结合了卷积神经的优点用于视觉特征提取的网络和用于处理有问题的结构。Xue W等[61]提出通过单元格之间的关系将表图像转换为其语法表示形式。为了重建表的句法结构，作者构建了一个单元格关系网络，在四个方向上预测每个单元格的相邻元素。在训练阶段，提出了基于距离的样本权值来处理类不平衡问题。根据检测到的关系，表用加权图表示，然后使用加权图来推断基本的语法表结构。

为了识别表格中的内部结构，例如复杂的表包含至少占用两列或两行的生成单元格。Chi Z[95]等提出了GraphTSR，将注意力模块集成到图网络中，并将图边缘分为水平、垂直和不相关，通过图神经网络来识别可移植文档格式(Portable Document Format，PDF)中的表结构。具体来说，它将表单元格作为输入，然后通过预测单元格之间的关系来识别表结构。类似得，在处理全局行和列信息，以及同时跨越单元格的问题上时，Lee E等[110]提出了基于网格形状图的表格识别方法，并提出了网格定位和网格元素分组网络。旨在利用网格结构并处理生成单元。

E Koci等[65]提出了一种采用图的方法表示图表内容的模型。鉴于此识别过程可以公式化为搜索输入图的最佳划分，其中图表中的与图纸的表格完全对应。作者定义一个客观的函数，以确定候选表格划分的优点。此功能经过调整以匹配特性给定数据集的。最后，作者使用遗传算法以搜索全局最优划分。Wenyuan Xue等[81]提出了TGRNet，由于各种不同的表格布局和样式。之前的方法通常将表建模为不同表单元格之间的标记序列或邻接矩阵，而没有解决表单元格逻辑位置的关系。本文将表结构识别问题重新表述为表图重构问题，提出了一个用于表结构识别的端到端可训练表格图重构网络。具体来说，该方法有两个主要分支，即单元格检测分支和单元格逻辑定位分支，共同预测不同单元格的空间位置和逻辑位置。

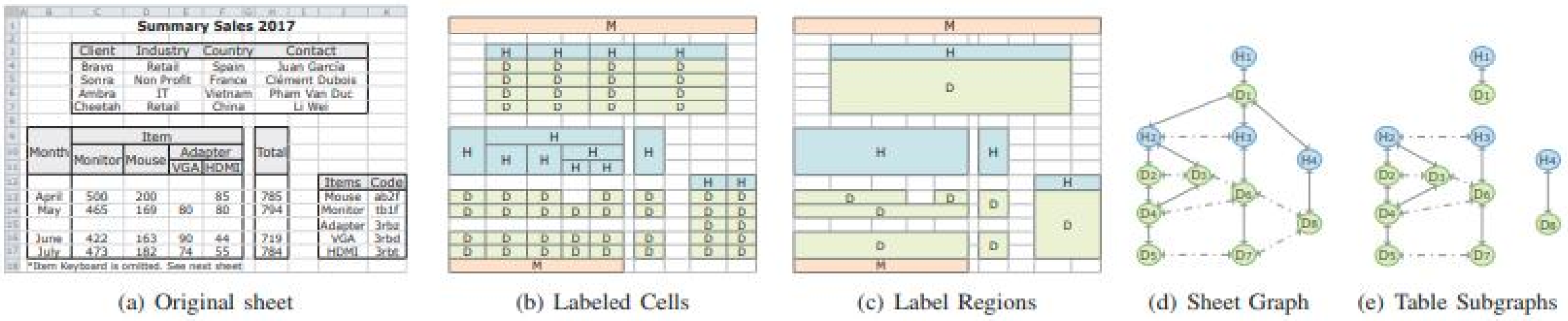

**图 10 基于图模型方法[65]**

**Fig. 10 Based on graph model method[65]**

#### 2.3.2.4 图像到标记序列方法

图像到标记序列方法借鉴了自然语言处理的思想，将问题转换为序列生成任务，因此需要内部表结构表示语言，通常使用标准标记语言(例如HTML，LateX，Markdown)实现。理论上，图像到标记序列方法相对于基于目标检测和基于图模型方法具有直接预测表结构的天然优势。因此，不需要后处理或规则来获得表结构，这是基于目标检测和基于图模型方法所必需的。但是在真正的实验中，这并不完全正确，因为预测的表结构标记序列不一定是语法正确的。因此，根据预测序列的质量，需要执行一些后处理以确保语法上有效(更不用说正确)的序列。

Deng Y等[64]将一种流行的注意力编码器-解码器（EDD）模型应用于表格结构识别中，并展示了端到端神经网络的潜力，以及相关的挑战。在高层次上，该模型由一个编码器部分和一个解码器部分组成，前者将输入图像表示为一组特定位置的特征，后者根据编码器的特征产生标签序列(即代码)。Kayal P.等[111]采用基于转换的语言建模范式进行表格的结构和内容提取。具体来说，建议的模型将表格图像转换为相应的LateX源代码。J. Li等[92]提出了一种自监督预训练的文档图像Transformer模型，该模型使用大规模未标记的文本图像进行文档分析，包括表格检测。由于缺乏人工标记的文档图像，没有监督的对应对象存在。作者利用DiT作为各种基于视觉的文档人工智能任务

的骨干网络，包括文档图像分类、文档布局分析、表检测以及OCR的文本检测。

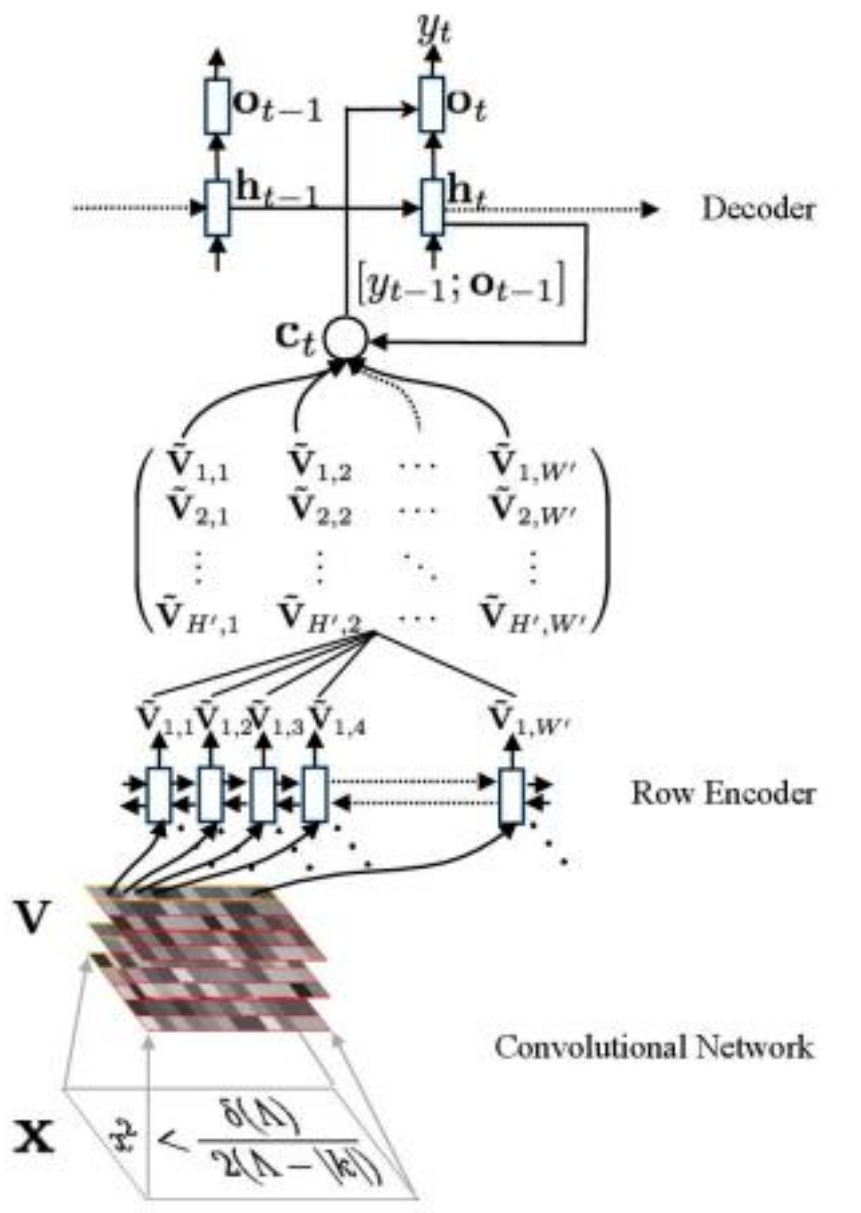

**图 11 注意力编码器-解码器（EDD）模型[64]**

**Fig. 11 Attention encoder-decoder (EDD) model[64]**

SA Khan等[63]在面对多变和噪声因素较大的表结构时，提出了基于深度学习的鲁棒解决方案，用于从文档图片中的可识别表中提取行和列。使用门控循环单元(Gated Recurrent Unit，GRU)和带有Softmax激活的全连接层将表格图片发送到双向循环神经网络之前，对其进行预处理。模型从上到下以及从左到右扫描图像，并对每个进行分类作为行分隔符或列分隔符输入。

Zhong X等[30]提出了一种新的基于注意力的编码器-双解码器(EDD)架构，将表格图像转换为HTML代码。该模型具有结构解码器，可以重构表结构，帮助单元格解码器识别单元格内容。此外，作者还提出了一种新的基于树编辑距离(TEDS)用于表识别的度量，它比预先建立的度量更合适地捕获跨越多行的单元格偏差和OCR错误。在此基础上，Ahmed Nassar等[95]提出一种基于Transformer的新的表结构识别模型TableFormer。模型改进了最新的端到端深度学习模型(即编码器-双解码器)。首先，作者引入了一种新的单元格对象检测解码器。这样，我们就可以直接从可编程可移植文档格式（Portable Document Format，PDF）源中获取表格单元格的内容，避免了自定义OCR解码器的训练。该模型用基于Transformer的解码器替换LSTM解码器，显著提高了之前最先进的树编辑距离分数(TEDS)。

在处理复杂表结构的问题上，B Xiao等[67]提出CATT-Net，一个复杂的表格结构可以用一个图来表示，其中顶点和边代表单个单元格以及它们之间的连接。然后，作者设计了一个条件注意力网络，并将表结构识别问题表征为一个单元格联分类问题。Yibo Li等[82]模型包含一个卷积编码器和两个并行行列解码器。编码器利用卷积块提取视觉特征;解码器将特征映射形成序列，并使用序列标记模型双向长短期记忆网络(BiLSTM)检测行、列分隔符。

Hangdi Xing等[94]提出了LORE模型。作者给表格结构识别提出了另一种范式，将表格结构识别作为一个逻辑位置回归问题，首次将逻辑回归和表格结构识别结合起来，区别于之前的方法通过预测检测到的单元格框的邻接关系或学习从表图像中生成相应的标记序列来解决这个问题。然而，它们要么依赖额外的启发式规则来恢复表结构，要么需要大量的训练数据和耗时的顺序解码器。

在图像到标记序列方法中，Lysak M.等[113]从图像到序列转化表示层面提出了一种优化表格结构表示语言(OTSL)，具有最小化的词汇和特定的规则。OTSL将令牌的数量减少到5个(HTML需要28个以上)，并将序列长度平均缩短到HTML的一半。因此，模型精度显著提高，与基于html的模型相比，推理时间缩短了一半，预测的表格结构在语法上总是正确率也大幅提高。这反过来又消除了大多数后处理需求。

| 表格检测 | | 表格结构识别 | |
| --- | --- | --- | --- |
| 目标检测方法 | Hao[8], Yi[96]<br>N. Sun[16], A. Gilani[13]<br>Y Huang[17], Madhav Agarwal[79]<br>Saman Arif[102]<br>Siddiqui[58], Tahira[112] | 目标检测方法 | SA Siddiqui[58]<br>X Zheng[70]<br>R. Long[107] |
| | | 语义分割方法 | Sebastian Schreiber[4]<br>SS Paliwal[54], SA Siddiqui[58]<br>Z. Zhang[71], C. Tensmeyer[62]<br>Qiao L[108], Hongyi Wang[114] |
| 语义分割方法 | Yang[105]<br>He[106]<br>I Kavasidis[52] | 图模型方法 | E Koci[57], SR Qasim[60]<br>Xue W[61],Chi, Z[95]<br>Lee, E[110]<br>E Koci[65], Wenyuan Xue[81] |
| | | 图像到标记序列方法 | Deng, Y[64], Kayal P.[111]<br>J. Li[92], SA Khan[63]<br>Zhong X[30], Ahmed Nassar[95]<br>B Xiao[67], Yibo Li[82]<br>Hangdi Xing[94], Lysak M.[113] |

**表 2 具有代表性的模型分类**

**Table. 2 Classification of representative models**

## 3 端到端（End-to-End）方法

表格识别领域，大部分工作专注于非端到端方法，也即将问题分为两个独立的子问题:表格检测以及表结构识别，然后尝试使用两个独立的方法解决每个子问题。也有一些研究者尝试使用端到端的方法来一次性解决表格识别问题，即输入是原始的图片或者电子文档等，输出就是处理好的表格形式。这样的方法不可避免会带来更多的困难，但是很多研究者依旧在这个方法上取得了优异的效果，本综述挑选了其中具有代表性的工作进行阐述说明。

Sebastian Schreiber等[4]于2017年提出的DeepDeSRT是首次采用Faster R-CNN对表检测和结构识别的端到端方法，随后，基于深度学习的语义分割进行表结构识别。基于规则式的方法依赖于启发式方法或附加的可移植文档格式（Portable Document Format，PDF）元数据(例如，打印指令、字符边界框或线段)，与之相反，DeepDeSRT是数据驱动的，不需要任何启发式方法或元数据来检测和识别文档图像中的表格结构，这个想法可以说是完全从深度学习角度出发的。

Madhav Agarwal等[79]于2020年提出了CDeC-Net模型，该模型由Mask R-CNN的多级扩展组成，具有可变形卷积的双主干，用于检测尺度变化的表，在更高的IoU阈值下具有较高的检测精度。作者通过大量实验在公开可用的基准数据集上对CDeC-Net进行了实证评估，都取得了很有竞争力的结果。

W Xue等[61]于2019年提出了ReS2TIM模型，提出了一个新的框架来转换表格图像通过其单元格之间的关系转化为其句法表示。为了重构句法，在表的结构中，作者构建了一个单元格关系网络，预测每个表格在四个方向上的拐角。在训练阶段，提出了一种基于距离的样本权重处理不平衡的问题。根据检测到的关系，该表由加权图表示然后用来推断基本的表格结构。

C Tensmeyer等[62]于2019年提出了SPLERGE （Split and Merge）模型，一种使用扩张卷积的方法。作者的方法需要使用两个不同的深度学习模型，第一个模型建立表格的网格状布局，第二个模型确定是否可以进一步跨越许多行或列的单元格。该模型也在ICDAR 2013可移植文档格式（Portable Document Format，PDF）表格竞赛数据集上获得了最先进的结果。

Y Deng等[64]于2019年研究了端到端表识别目前存在的问题，作者通过构建由450K表格图像组成的大型数据集与相应的LaTeX源配对。应用流行的注意力编码器-解码器模型到该数据集并显示端到端训练的神经系统。

SS Paliwal等[54]于2020年提出了TableNet模型，这是一种用于表格检测和结构识别的新型端到端深度学习模型。作者添加额外的空间语义训练，进一步提升了模型表现。为了划分表格和列区域，该模型利用了表格检测和表格结构识别这两个孪生目标之间的依赖关系。然后，从发现的表格子区域中，执行基于语义规则的行提取。该模型微调后在其他数据集上依旧表现很好，适合迁移学习。该模型在ICDAR 2013和Marmot数据集上都取得了优异的效果。

Devashish Prasad等[53]于2020年提出了CascadeTabNet模型，一种基于Cascade Mask R-CNN模型的端到端深度学习方法。其中R-CNN HRNet模型用于表检测和结构识别，作者提出了一种改进的基于深度学习的终端，使用单个卷积同时解决表格检测和表格结构识别问题的端到端方法。该模型在2019年ICDAR竞赛中成绩中排名第三。

Danish Nazir等[80]于2021年提出了HybridTabNet，一种端到端可训练管道，用于扫描文档图像中的表格

检测。两阶段表格检测器使用ResNeXt-101主干进行特征提取，并使用混合任务级联来定位扫描文档图像中的表。此外，作者在骨干网络中用可变形卷积代替了传统的卷积。该模型在ICDAR 2013，ICDAR 2017，ICDAR 2019，TableBank，Marmot，UNLV数据集上都超越了之前最先进的结果（State of the art）。

Hashmi K.A等[83]于2021年提出了CasTabDetectoRS模型，作者在Cascade Mask R-CNN上训练，包括现有主干架构中的递归特征金字塔网络和可切换的亚历斯卷积。通过使用相对轻量级的ResNet-50骨干网络提取特征，该模型在不依赖于预处理和后处理方法、较重的骨干网（ResNet-101、ResNeXt-152）和内存密集型可变形卷积的情况下，可以获得更好的结果。该模型在ICDAR 2019，TableBank， UNLV和Marmot数据集上都取得了超越之前最先进方法的结果（State of the art）。

Wenyuan Xue等[81]于2021年提出了TGRNet模型，由于各种不同的表格布局和样式。之前的方法通常将表建模为不同表单元格之间的标记序列或邻接矩阵，而没有解决表单元格逻辑位置的关系。本文将表结构识别问题重新表述为表图重构问题，提出了一个用于表结构识别的端到端可训练表格图重构网络(TGRNet)。具体来说，该方法有两个主要分支，即单元格检测分支和单元格逻辑定位分支，共同预测不同单元格的空间位置和逻辑位置。

A Jain等[74]于2022年通过训练一个深度网络来识别表格图片中包含的各种单词对 之间的空间关系，以便识别表格结构。作者提供了一个端到端的管道，称为TSR- dsaw：通过深空间关联的词，它生成一个数字表示的表格图片在一个结构化的格式，如HTML。该技术首先利用一个文本检测网络，如CRAFT，来识别输入表格图片中的每个单词。接下来，使用动态规划，创建单词对。这些单词对在每个单独的图像中加下划线，然后交给DenseNet-121分类器，该分类器经过训练，可以识别空间相关性，如同行、同列、同单元或无。最后，作者对分类器输出进行后处理，以生成HTML表格结构。模型在PubTabNet和ICDAR 2013数据集上，演示改进最先进方法的结果（State of the art）。

Ly N T等[77]于2023年提出了一个弱监督模型WSTabNet用于表识别问题，该模型仅依赖于表图像的HTML(或LaTeX)代码级注释。该模型由三个主要部分组成:用于特征提取的编码器、用于生成表结构的结构解码器和用于预测表中每个单元的内容的单元解码器。同时通过随机梯度下降算法进行端到端的训练，只需要表图像及其基本真实的HTML(或LaTeX)表示。该模型解决了以前的大多数表识别方法依赖于包含许多注释的表格图像的训练数据集的问题。

Ly N T等[84]于2023年提出了一种基于图像的表识别的端到端多任务学习模型。该模型由一个共享编码器、一个共享解码器和三个独立的解码器组成，用于学习表识别的三个子任务:表结构识别、单元检测和单元内容识别。整个系统可以很容易地以端到端的方式进行训练和推断。作者在FinTab和PubTabNet上评估了所提出模型的性能。实验结果表明，该模型在所有基准数据中都优于目前最先进的方法（State of the art）

| Literature | Model | Year | Method |
|---|---|---|---|
| Sebastian Schreiber [4] | DeepDeSRT | 2017 | 1. 采用Faster R-CNN对表检测和结构识别。<br>2. 基于深度学习的语义分割进行表结构识别。 |
| Madhav Agarwal [79] | CDeC-Net | 2020 | 1. 由Mask R-CNN的多级扩展组成。<br>2. 具有可变形卷积的双主干，用于检测尺度变化的表格。 |
| W Xue[61] | ReS2TIM | 2019 | 1. 转换表格图像通过其单元格之间的关系转化为其句法表示。<br>2. 构建了一个单元格关系网络，预测每个表格在四个方向上的拐角。<br>3. 根据检测到的关系，由加权图表示然后用来推断基本的表格结构。 |
| C Tensmeyer[62] | SPLERGE | 2019 | 1. 使用扩张卷积的方法，一个模型建立表格的网格状布局。<br>2. 另外一个模型确定是否可以进一步跨越许多行或列的单元格。 |
| Y Deng[64] | TABLE2LATEX | 2019 | 1. 构建由450K表格图像组成的大型数据集与相应的LaTeX源配对<br>2. 应用注意力编码器-解码器模型。 |
| SS Paliwal[54] | TableNet | 2020 | 1. 添加额外的空间语义训练。<br>2. 利用了表格检测和表格结构识别这两个孪生目标之间的依赖关系。<br>3. 执行基于语义规则的行提取。 |
| Devashish Prasad[53] | CascadeTabNet | 2020 | 1.单个卷积同时解决表格检测和表格结构识别问题的端到端方法。 |
| Danish Nazir[80] | HybridTabNet | 2021 | 1. 使用ResNeXt-101主干进行特征提取，并使用混合任务级联(HTC)来定位扫描文档图像中的表格。<br>2. 在骨干网络中用可变形卷积代替了传统的卷积。 |
| Hashmi K.A[83] | CasTabDetectoRS | 2021 | 1. 递归特征金字塔网络和可切换的亚历斯卷积。<br>2. 使用相对轻量级的ResNet-50骨干网络提取特征。 |
| Wenyuan Xue[81] | TGRNet | 2021 | 1. 将表结构识别问题重新表述为表图重构问题。<br>2. 单元格检测分支和单元格逻辑定位分支，共同预测不同单元格的空间位置和逻辑位置。 |
| A Jain[74] | TSR- dsaw | 2022 | 1. 生成一个数字表示的表格图片在一个结构化的格式<br>2. 利用一个文本检测网络来识别输入表格图片中的每个单词。<br>3. 使用动态规划，创建单词对。这些单词对在每个单独的图像中加下划线，然后交给分类器，对分类器输出进行后处理，以生成表格结构 |
| Ly N T[77] | WSTabNet | 2023 | 1. 用于特征提取的编码器、用于生成表结构的结构解码器和用于预测表中每个单元的内容的单元解码器。<br>2. 通过随机梯度下降算法进行端到端的训练。 |
| Ly N T[84] | Multi-Task | 2023 | 1. 一个共享编码器、一个共享解码器和三个独立的解码器。 |

**表 3 代表性的端到端方法**

**Table 3 Representative end-to-end methods**

## 4 以数据为中心(Data-Centric)方法

本文上面我们从表格检测和结构识别两个子任务的模型以及端到端的模型回顾了表格识别领域的发展。但是随着深度学习的不断演进发展，我们注意到除了以模型为中心的方法以求解决表格识别问题之外，还有研究者引入了以数据为中心的方法（Data-Centric），例如数据的预处理和后处理、数据增强等方法，都在这个领域取得了一些不可忽视的成果。

在传统的以模型为中心的人工智能生命周期中，研究人员和开发人员主要关注识别更有效的模型，以提高人工智能的性能，同时保持数据基本不变。然而，这种以模型为中心的范式忽略了数据的潜在质量问题和不希望看到的缺陷，比如缺失的值、不正确的标签和异常。以数据为中心的人工智能强调对数据进行系统工程来构建人工智能系统，将我们的重点从模型转移到数据[91]。

研究者在表格处理领域以数据为中心的方法目前并不多，只有在零星几篇文章有予以说明解释，尽管如此，以数据为中心的方法往往都能解决一些领域内的顽疾，很有启发意义，因而本文从实例分割、数据预处理与后处理、数据增强、基准对齐四个方面各自选取了一篇论文进行阐述说明，一方面介绍这些工作的同时，也希望给后来的研究者带来一些思考。

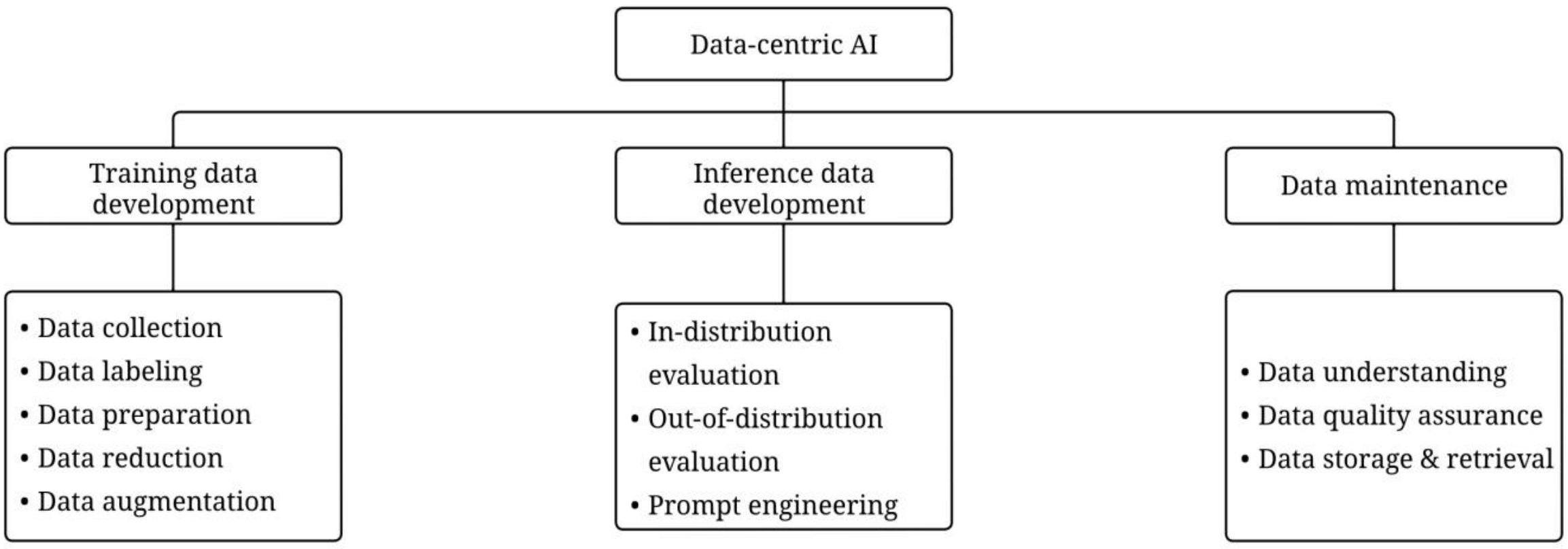

**图 12 以数据为中心方法示意图[91]**

**Fig. 12 Diagram of a Data-Centric method[91]**

### 4.1 实例分割

Li XH等[85]于2021年提出了一种基于实例分割的表格结构识别方法，该方法利用了单元格内容和单元格框线信息。为了克服表格不一致的影响，作者设计了一种基于每个文档图像中平均单元格大小和标尺线密度的自适应图像缩放方法。在这样的方法下，该方法自适应缩放将每个图像的大小调整到一个最优的大小，不仅可以提高整体模型的性能，而且可以平均减少内存和计算开销。

### 4.2 预处理与后处理

Mikhailov和A. Shigarov[86]于2021年提出使用预处理算法来识别可移植文档格式（Portable Document Format，PDF）页面中的标题、段落和图像。这允许在页面中选择可以放置真正的表格所在的区域。然后我们使用深度神经网络来预测这些区域的表。作者还

提出了后处理算法来验证预测，并在表检测后过滤掉错误的候选表。这样解决了可移植文档格式（Portable Document Format，PDF）表检测预测页面上表的边界框，有些预测不可避免地是错误的问题。

### 4.3 数据增强

Khan U等[87]于2021年提出了TabAug方法，这是一种重新构图（re-imagined）的数据增强技术，通过复制和删除行和列来产生表格图像的结构变化。它还包含一个数据驱动的概率模型，该模型允许对增强过程进行控制。这种方法解决了这个领域的大量数据集仍然不可用，而注释新数据既昂贵又耗时的问题。

### 4.4 基准对齐

Smock B.等[88]于2023年提出必须仔细处理用于表格结构识别的基准数据集，以确保它们得到一致的注释。然而，即使数据集的注释是自一致的，数据集之间也可能存在显著的不一致，这可能会损害在这些数据集上训练和评估的模型的性能。作者证明了对齐这些基准，消除它们之间的错误和不一致，可以显著提高模型的性能。作者采用表格转换器在整个方法的过程中，通过一种以数据为中心的方法来展示这一点。

## 5 实验结果

### 5.1 表格检测实验结果

表格检测任务是通过模型或者算法识别出表格所在的区域并回归指定为表格区域的边界框。评估表格检测的过程中，召回率（Recall），精度值（Precision），F1指标（F1-Measure）是表格检测任务中最常用的指标，可以很好地分析模型在表格检测任务上的效果。其中ICDAR 2013、ICDAR 2017、ICDAR 2019和UNLV等数据集通常用于评估表检测算法的有效性。此外在实验结果中，交集面积/联合面积(Intersection over Union，IoU)值也至关重要，设定较高的阈值往往会导致模型精度等指标的降低，因而本文也将表格检测中阈值的设置也展示出来，以供比较。值得注意的是，有几种方法没有指定IoU阈值，而是将他们的发现与其他方法进行了比较。因此，对于这些方法，我们考虑了相同的阈值。

| Literature | Dataset | Method | IoU | Recall | Precision | F1-Measure |
|---|---|---|---|---|---|---|
| Hao[8] | Private | Object-detection | | 0.9724 | 0.9215 | 0.9463 |
| Yi[96] | Private | Object-detection | 0.8 | 0.832 | 0.815 | |
| N. Sun[16] | ICDAR 2017 | Object-detection | 0.6 | 0.956 | 0.943 | 0.949 |
| A. Gilani[13] | UNLV | Object-detection | | 0.9067 | 0.823 | 0.8629 |
| Y Huang[17] | ICDAR 2013 | Object-detection | 0.6 | 0.936 | 0.986 | 0.961 |
| Y Huang[17] | ICDAR 2013 | Object-detection | 0.8 | 0.846 | 0.892 | 0.868 |
| Y Huang[17] | ICDAR 2017 | Object-detection | 0.6 | 0.972 | 0.978 | 0.975 |
| Y Huang[17] | ICDAR 2017 | Object-detection | 0.8 | 0.968 | 0.975 | 0.971 |
| Madhav Agarwal[79] | ICDAR 2013 | Object-detection | 0.9 | 0.660 | 0.660 | 0.660 |
| Madhav Agarwal[79] | ICADR 2017 | Object-detection | 0.9 | 0.902 | 0.957 | 0.929 |
| Madhav Agarwal[79] | ICADR 2019 | Object-detection | 0.9 | 0.895 | 0.934 | 0.915 |
| Madhav Agarwal[79] | UNLV | Object-detection | 0.9 | 0.496 | 0.618 | 0.557 |
| Madhav Agarwal[79] | Marmot | Object-detection | 0.9 | 0.823 | 0.891 | 0.857 |
| Madhav Agarwal[79] | TableBank | Object-detection | 0.9 | 0.966 | 0.982 | 0.974 |
| Madhav Agarwal[79] | PubLayNet | Object-detection | 0.9 | 0.965 | 0.983 | 0.974 |
| Saman Arif[102] | UNLV | Object-detection | | 0.9321 | 0.8633 | 0.8964 |
| Siddiqui[58] | TabStructDB | Object-detection | | 0.9308 | 0.9319 | 0.9298 |
| Tahira[112] | ICDAR 2019 | Object-detection | 0.8 | 0.921 | 0.949 | 0.935 |
| Tahira[112] | ICDAR 2019 | Object-detection | 0.9 | 0.913 | 0.926 | 0.919 |
| He[106] | Marmot-chin | Semantic-segmentation | 0.8 | 0.761 | 0.77 | |
| He[106] | Marmot-chin | Semantic-segmentation | 0.9 | 0.491 | 0.493 | |
| He[106] | Marmot-eng | Semantic-segmentation | 0.8 | 0.70 | 0.753 | |
| He[106] | Marmot-eng | Semantic-segmentation | 0.9 | 0.45 | 0.47 | |
| I Kavasidis[52] | ICDAR 2013 | Semantic-segmentation | 0.5 | 0.981 | 0.975 | 0.978 |

**表 4 表格检测实验结果**

**Table 4 Experimental result on table detection**

根据表4的实验数据来看，基于目标检测方法的表格检测被大多数研究者所采用，模型取得的效果相较于语义分割方法也好一些。

## 5.2 表格结构识别实验结果

表格结构识别任务是通过模型或者算法在已经检测好的表格的基础上识别和解释表格的布局和组织。其中包括表格行、表格列、表格中的单元格等一系列元素。表格结构识别的目标在于准确识别文档中的所有表，即使它们具有不同的样式或格式。在表格结构识别任务中召回率（Recall），精度值（Precision），F1指标（F1-Measure）依旧表格结构识别任务中最常用的指标，受到大部分研究者的使用。但是这些常用指标存在着对跨行表格结构评估不清晰已经对后续信息识别评估不佳等问题。因而树-编辑-距离的相似度（Tree-Edit-Distance-based Similarity，TEDS）也得到了一些研究者的注意，尤其是采用图像到标记序列方法的研究者们。本文将单独将基于树-编辑-距离的相似度评价指标的模型或者算法也呈现出来，以供参考。数据集方面，ICDAR 2013、ICDAR 2019还是最常用的评估数据集，此外其他数据集都有一定使用。值得一提的是，与表格检测不同，表格结构识别任务中很多方法没有指定IoU阈值，因此对于IoU阈值这一项本文也未给出。

| Literature | Dataset | Method | Recall | Precision | F1-Measure |
|---|---|---|---|---|---|
| SA Siddiqui[58] | ICDAR 2013 | Object-detection | 0.930 | 0.931 | 0.930 |
| X Zheng[70] | ICDAR 2013 | Object-detection | 0.927 | 0.944 | 0.935 |
| R. Long[107] | ICDAR 2019 | Object-detection | 0.883 | 0.955 | 0.917 |
| Sebastian Schreiber[4] | Private | Semantic-segmentation | 0.874 | 0.959 | 0.914 |
| SS Paliwal[54] | Marmot | Semantic-segmentation | 0.8987 | 0.9215 | 0.9098 |
| Z. Zhang[71] | SciTSR | Semantic-segmentation | 0.9415 | 0.9669 | 0.9540 |
| C. Tensmeyer[62] | Private | Semantic-segmentation | 0.866 | 0.869 | 0.868 |
| Qiao L[108] | ICDAR2013 | Semantic-segmentation | 0.991 | 0.967 | 0.979 |
| Hongyi Wang[114] | ICDAR2013 | Semantic-segmentation | 0.956 | 0.945 | 0.950 |
| E Koci[57] | UNLV | Graph | 0.367 | 0.963 | |
| SR Qasim[60] | Synthetic 500k | Graph | 0.934 | 0.934 | 0.934 |
| Xue W[61] | ICDAR 2013 | Graph | 0.747 | 0.734 | 0.740 |
| Chi Z[95] | ICDAR 2013 | Graph | 0.747 | 0.734 | |
| Lee E[110] | SciTSR | Graph | 0.7681 | 0.8920 | 0.8038 |
| Xue W[81] | ICDAR 2019 | Graph | 0.798 | 0.860 | 0.828 |
| J. Li[92] | Private | Image-to-markup-sequence | 0.9317 | 0.9543 | 0.9429 |
| SA Khan[63] | ICDAR 2013 | Image-to-markup-sequence | 90.12 | 96.92 | 93.39 |
| B Xiao[67] | SciTSR | Image-to-markup-sequence | 0.981 | 0.981 | 0.981 |
| B Xiao[67] | SciTSR-COMP | Image-to-markup-sequence | 0.982 | 0.982 | 0.982 |
| B Xiao[67] | ICDAR2013 | Image-to-markup-sequence | 0.963 | 0.963 | 0.963 |
| Hangdi Xing[94] | ICDAR2013 | Image-to-markup-sequence | 0.986 | 0.992 | 0.989 |
| Hangdi Xing[94] | SciTSR-comp | Image-to-markup-sequence | 0.992 | 0.994 | 0.993 |
| Hangdi Xing[94] | ICDAR 2019 | Image-to-markup-sequence | 0.887 | 0.879 | 0.883 |
| Hangdi Xing[94] | WTW | Image-to-markup-sequence | 0.945 | 0.959 | 0.951 |

**表 5 表格结构识别实验结果**

**Table 5 Experimental result on table structure recognition**

| Model | TEDS | | |
|---|---|---|---|
| | Simple | Complex | All |
| Hongyi Wang[114] | | | 90.7 |
| C-CenterNet[107] | | | 83.3 |
| EDD[30] | 91.2 | 85.4 | 88.3 |
| TableFormer[95] | 95.4 | 90.1 | 93.6 |
| LGPMA[108] | | | 94.6 |
| GTE[70] | | | 93.0 |
| SEM[71] | 94.8 | 92.5 | 93.7 |

**表 6 表格结构识别 TEDS**

**Table 6 TEDS on table structure recognition**

综合表5和表6的实验数据来看，脱胎于自然语言处理领域的图像到标记序列方法在表格结构识别任务上展示出了独特的优势，无论是传统评价指标还是TEDS这种专注于表格结构某些特点的指标。但是其他方法依旧有研究者在进行探索。

## 6 总结与展望

表格识别领域的问题近几年已经吸引了大量学者的研究，很难逐一分析每一位学者的研究，因而本文选择了比较有代表性的方法进行分析。本文通过公开数据集、评价指标；表格识别领域代表性的模型；端到端方法代表性的模型；以数据为中心方法代表性的模型以及表格识别领域实验数据的分析来对表格识别这一领域进行了全面的梳理。

从结果来看，经过深度学习技术的引入和运用，表格检测这一任务目前随着计算机视觉的快速发展，研究者从目标检测的思想中汲取灵感，改进方法，在各个公开数据集上取得了很好的效果，也不乏经过一定微调泛用性极好的模型。但是调高IoU的阈值，各个方法的指标都会有一定下降，这也意味着我们还有不足需要弥补。对于表格结构识别来说，其中基于图模型的方法和基于图像到标记序列的方法更受关注一些，其中基于图像到标记序列的方法在更多数据集中取得了优越的结果，但是仍需要更多创造性的工作去完成表格结构识别这一具有挑战性的课题。

此外端到端的方法以及以数据为中心的方法与主流的表格检测、表格结构识别相比并未有太多的关注，很值得研究者的进一步投入与研究。可以预见的是，一旦表格被精准识别理解，结构化地分割表格然后识别内容的工作就随着而来了。

除此之外，我们还能看到不断有更新，更详尽，更标准的数据集的引入并得到研究者们的普遍认可。也能看到有新的标准的提出并被采纳。这也意味着表识别理及其衍生出的表格检测、表格结构识别、端到端、数据为中心这些子领域的问题仍然生生不息，值得更多的研究和投入。

## 参 考 文 献